%% file: arxiv.tex
\documentclass[10pt,twocolumn,letterpaper]{article}
\pdfoutput=1

\usepackage{iccv}
\usepackage{times}
\usepackage{epsfig}
\usepackage{graphicx}
\usepackage{amsmath}
\usepackage{amssymb}

\usepackage{csquotes} 
\usepackage{diagbox}
\usepackage{multirow}
\usepackage{multicol}
\usepackage{bbm}
\usepackage{algorithm}
\usepackage{algpseudocode}
\usepackage[dvipsnames]{xcolor}
\usepackage[normalem]{ulem}
\usepackage{subfig}
\usepackage{etoc}
\usepackage[accsupp]{axessibility}  

\usepackage[breaklinks=true,bookmarks=false]{hyperref}

\iccvfinalcopy 

\def\iccvPaperID{10595} 

\ificcvfinal\fi

\begin{document}

\title{Homography Guided Temporal Fusion for Road Line and Marking Segmentation }

\author{Shan Wang$^{1,2}$ \quad Chuong Nguyen$^{1}$ \quad Jiawei Liu$^{2}$ \quad Kaihao Zhang$^{2}$ \quad  Wenhan Luo$^{3}$ \\ Yanhao Zhang$^{2}$ \quad Sundaram Muthu$^{1}$ \quad Fahira Afzal Maken$^{1}$ \quad Hongdong Li$^{2}$\\
{\tt\small$^{1}$Data61, CSIRO} \quad  {\tt\small $^{2}$Australian National University} \quad {\tt\small$^{3}$Sun Yat-sen University}
}

\maketitle
\ificcvfinal\thispagestyle{empty}\fi

\begin{abstract}
 Reliable segmentation of road lines and markings is critical to autonomous driving. Our work is motivated by the observations that road lines and markings are (1) frequently occluded in the presence of moving vehicles, shadow, and glare and (2) highly structured with low intra-class shape variance and overall high appearance consistency. To solve these issues, we propose a Homography Guided Fusion (HomoFusion) module to exploit temporally-adjacent video frames for complementary cues facilitating the correct classification of the partially occluded road lines or markings. To reduce computational complexity, a novel surface normal estimator is proposed to establish spatial correspondences between the sampled frames, allowing the HomoFusion module to perform a pixel-to-pixel attention mechanism in updating the representation of the occluded road lines or markings. Experiments on ApolloScape, a large-scale lane mark segmentation dataset, and ApolloScape Night with artificial simulated night-time road conditions, demonstrate 
 that our method outperforms other existing SOTA lane mark segmentation models with less than 9\% of their parameters and computational complexity. We show that exploiting available camera intrinsic data and ground plane assumption for cross-frame correspondence can lead to a light-weight network with significantly improved performances in speed and accuracy. We also prove the versatility of our HomoFusion approach by applying it to the problem of water puddle segmentation and achieving SOTA performance \footnote{Code is available at \href{https://github.com/ShanWang-Shan/HomoFusion}{https://github.com/ShanWang-Shan/HomoFusion}.}.
\end{abstract}

\etocdepthtag.toc{mtchapter}
\etocsettagdepth{mtchapter}{subsection}
\etocsettagdepth{mtappendix}{none}

\vspace{-0.3cm}
\section{Introduction}
\label{sec:intro}
\begin{figure}[!htb]
  \centering
  \includegraphics[width=0.99\linewidth]{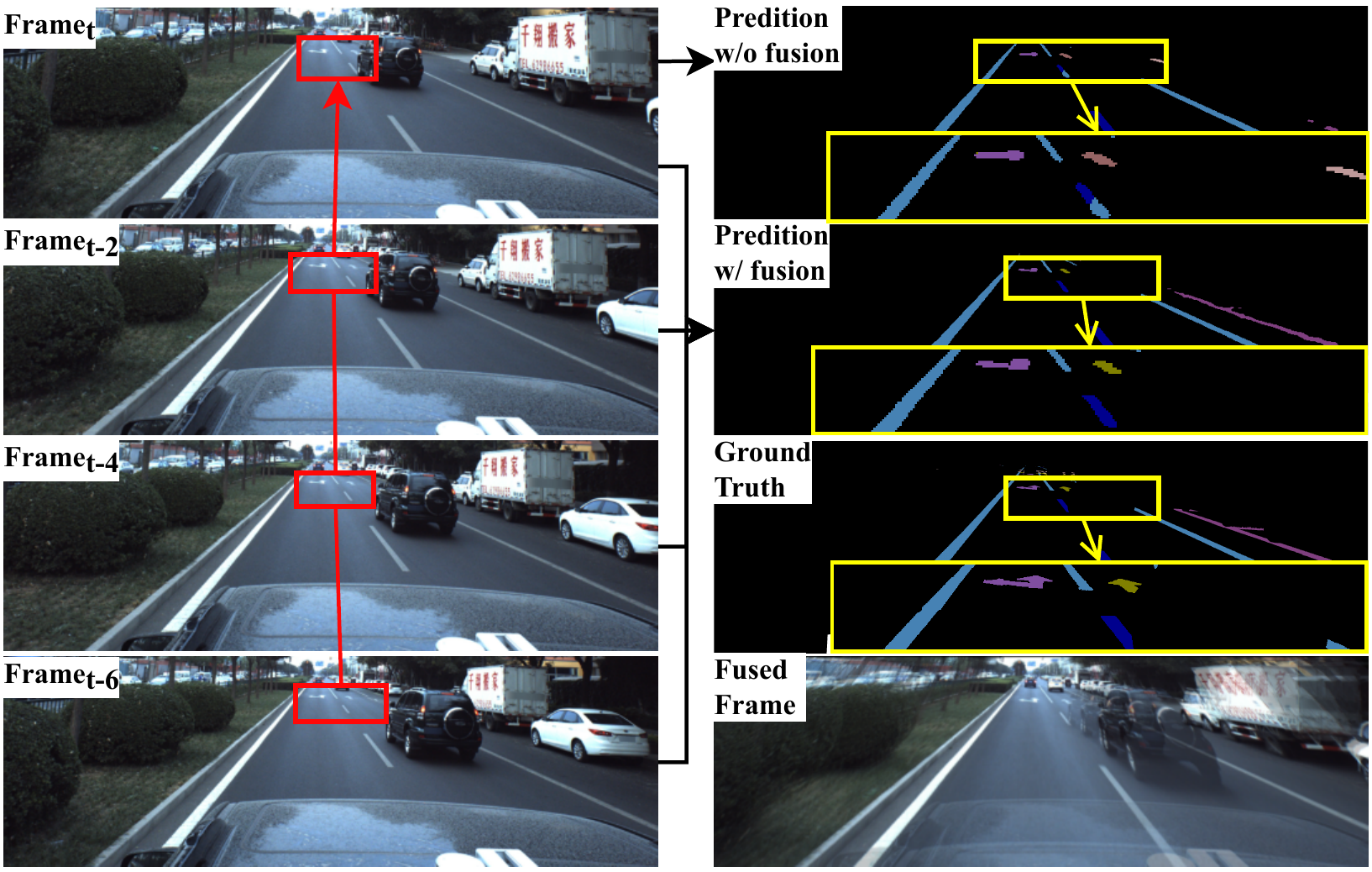}
  \caption{Illustration of the effect of the proposed HomoFusion module that explores the adjacent frames for cues, facilitating the correct classification of (1) a \textcolor{Green}{\enquote{Straight Arrow}}, which, with its bottom half occluded by a vehicle, is mistakenly classified as a \textcolor{Tan}{\enquote{Right Turn \& Straight Arrow}} without the HomoFusion module, and (2) a partially occluded \textcolor{Blue}{\enquote{Dotted Line}}, which is incorrectly classified as a \textcolor{CornflowerBlue}{\enquote{Solid Line}} without the HomoFusion module. 
  The fused frame in the $4^{th}$ row and the $2^{nd}$ column demonstrates the recovered road lines and markings after projecting the previous frames onto the current frame with the estimated homography matrices. The yellow box enlarges the area where mistake classifications are corrected. The red box indicates the spatially corresponding area across the frames. Best viewed in color.
  }
  \label{fig:fuse_impact}
  \vspace{-0.3cm}
\end{figure}

Lane mark segmentation aims to achieve pixel-wise classification of road lines and markings simultaneously. Known for its quintessential importance in autonomous driving, lane mark segmentation is also an effective tool 
for constructing accurate High Definition (HD) maps.
Existing works solve individual sub-tasks, \eg drivable area segmentation \cite{8594450,chen2019progressive} and lane detection \cite{9460822,8500547}, 
while few address the lane mark segmentation in its entirety \cite{yin2020fusionlane,hou2020inter}.

Despite the tremendous progresses in semantic/scene segmentation \cite{xie2021segformer,sun2022vss,liu2021learning}, little attention has been paid to the lane mark segmentation task \cite{yin2020fusionlane,hou2020inter}. Yin \etal \cite{yin2020fusionlane} leverages additional LiDAR information, merging the segmented visual information with the point clouds, to achieve lane mark segmentation. IntRA-KD \cite{hou2020inter} applies the knowledge distillation technique to improve the efficiency of the lane mark segmentation model.

A major challenge in the lane mark segmentation task is 
partial occlusion of the road lines and markings caused by the vehicle and surrounding environment leading to false classifications.
For example, a partially occluded straight arrow can be easily mistaken as a right turn \& straight arrow, as depicted in Fig.~\ref{fig:fuse_impact}. 
However, existing lane mark segmentation methods 
have not yet made use of the following observations:
(1) occlusion of the road lines and markings 
(due to nearby moving vehicles, shadow, and glare) can be reduced by cross-frame consistency; (2) road lines and markings are highly structured, maintaining high consistency in intra-class shapes as well as in overall appearance,
alleviating the requirements on learning complicated features to distinguish different classes with high intra-class variance as in semantic/scene segmentation.

To find complementary information from adjacent frames, we use the ground plane assumption for the road region immediately in front of the camera. Our Homography Guided Fusion (HomoFusion) module achieves temporally consistent representation, recovers partially occluded road lines or markings, and leads to correct classification. The module's success depends on accurate cross-frame spatial correspondence, which we achieve using a homography transformation matrix estimated with available intrinsic and extrinsic parameters of the on-vehicle camera, and the road surface normal estimated by our novel optimization method called Road Surface Normal Estimator (RSNE).

Using the highly structured nature of road lines and markings could reduce the complexity of the detection problem and make it suitable to run on edge devices.
To this end, we employ (1) a lightweight encoder to represent the visual information in feature spaces, and (2) a cross-frame pixel-to-pixel attention mechanism in our HomoFusion module, instead of a more computationally expensive global attention mechanism employed by other methods.
These design decisions allow the proposed model to outperform the existing lane mark segmentation models with less than 9\% of their parameters and Giga Floating Point Operations (GFLOPs)
We summarize our contributions as follows:
\begin{itemize}
  \vspace{-0.1cm}
    \item We propose a HomoFusion module that uses ground plane assumption and 
    the adjacent frames for temporally consistent representations for accurate classification of partially occluded road lines and markings.
      \vspace{-0.1cm}
    \item We design a novel estimator RSNE for road surface normal,
    which, combined with camera intrinsic and extrinsic parameters readily available on an autonomous vehicle, yields 
    accurate homography matrix between frame pairs. RSNE simplifies the 8 degrees of freedom (DoF) of homography problem to a 2 DoF of normal vector problem.
      \vspace{-0.1cm}
    \item We present a lightweight lane mark segmentation
    model that achieves better performance than the state-of-the-art (SOTA) methods with significantly reduced model complexity and computation requirements. 
\end{itemize}

\section{Related Work}
\subsection{Lane Detection}

Traditional lane detection approaches rely on handcrafted features such as color \cite{sun2006hsi,wang2014approach}, edge \cite{li2016nighttime} and texture \cite{li2016road}, which are limited 
in complex scenarios \cite{tsai2008lane,borkar2011novel}.
Recent advances in deep neural networks have led to significant performance improvements, with methods like message-passing networks \cite{pan2018spatial} and attention-based modules \cite{tabelini2021keep}.
\cite{hou2020inter,philion2019fastdraw,qin2020ultra,yoo2020end,liu2021condlanenet} formulate lane detection as a row-wise classification task based on grid division of the input image. 
PolyLaneNet \cite{tabelini2021polylanenet} is the first parametric prediction method, which outputs polynomials to represent each lane. 
BEVFormer \cite{li2022bevformer} and PETRv2 \cite{liu2022petrv2} employ transformer mechanisms to streamline the conversion of perspective views into bird's-eye views, thereby eliminating variations in object size caused by perspective.
Although the above methods achieve impressive performance, all of them focus on detecting the lane lines and ignore the basic road elements like arrow signal in Fig. \ref{fig:fuse_impact}. Furthermore, lane detection differs from segmentation tasks in that it focuses on identifying the boundaries of the drivable area, rather than the real shape of the lanes.

\begin{figure*}[!htb]
  \centering
  \begin{minipage}[b]{0.80\linewidth}
      \includegraphics[width=0.99\linewidth]{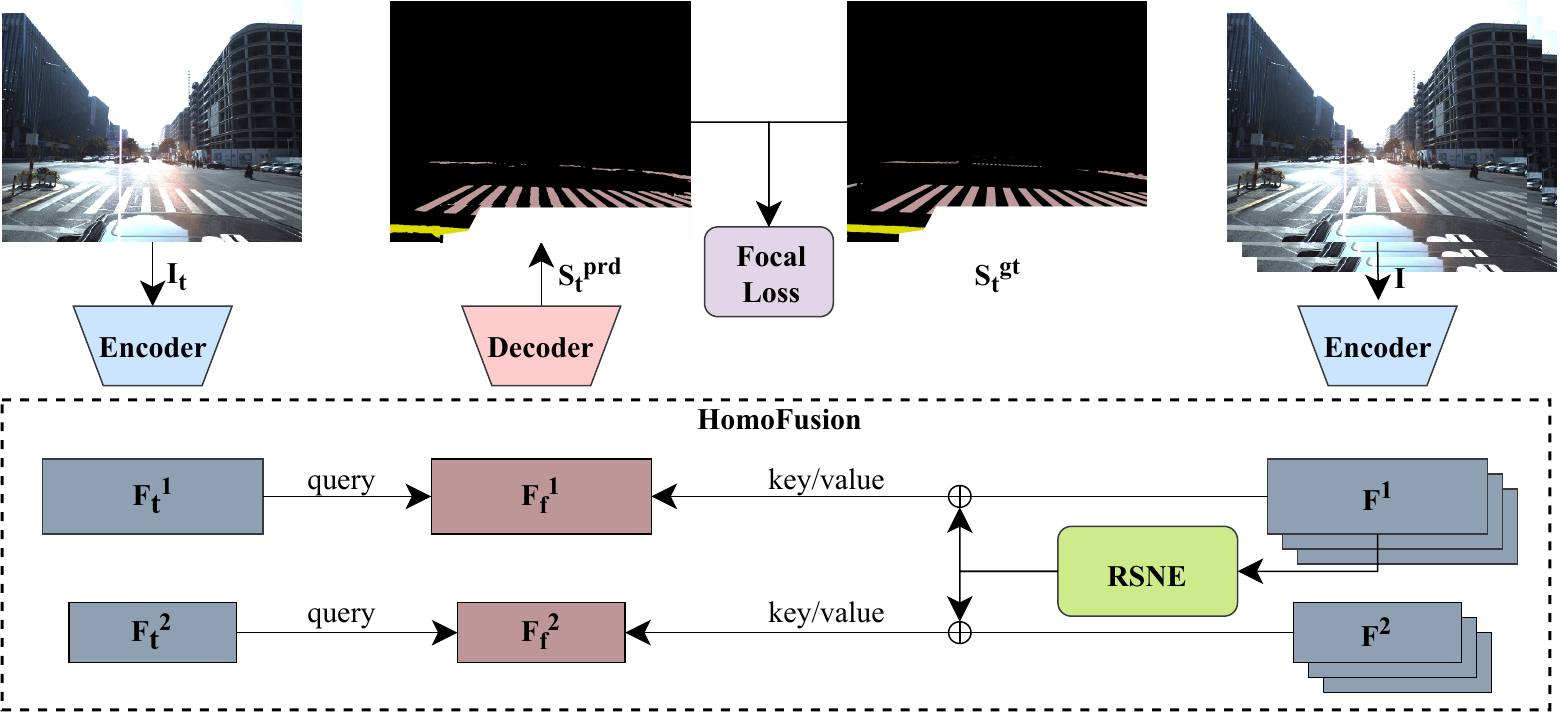}
  \end{minipage}
  \caption{Overview of our proposed model consisting of a pair of lightweight encoder and decoder, our proposed HomoFusion module, and our proposed Road Surface Normal Estimator (RSNE). A sequence of frames $\mathbf{I}$, including a target frame $\mathbf{I_{t}}$ and $n - 1$ previous frames, are encoded into the feature representations ($\mathbf{F}^{l}$). RNSE estimates the road surface normal vector, which, combined with the  camera intrinsic and extrinsic parameters, yields a homography matrix between each frame pair, establishing cross-frame spatial correspondences. HomoFusion uses pixel-to-pixel attention mechanism to obtain temporally consistent representation for on-road pixels of the current frame with the spatial correspondence across frames as guidance. Finally, the decoder decodes and upsamples the temporally consistent feature representations to produce the lane mark segmentation prediction ($\mathbf{S_{t}^{prd}}$).
  }
  \label{fig:architecture}
  \vspace{-0.5cm}
\end{figure*}

\subsection{Road Line and Marking Segmentation}
The identification of road elements is crucial for ensuring safety in autonomous driving systems, but there are 
few works addressing this problem.
Hou \textit{et al.} \cite{hou2020inter} 
introduced a distillation approach that demonstrates competitive performance in lane mark segmentation.
Yin \textit{et al.} \cite{yin2020fusionlane} used an LSTM-based network to segment images with DeepLabv3+ \cite{chen2018encoder} and then merged them with point clouds to assist with lane segmentation.
Unfortunately, these methods do not address the challenges of occlusion, shadows, and glare that frequently occur in real-world driving scenarios.
Recently, Zhang \textit{et al.} \cite{zhang2021vil} introduced global memory information from previous frames to enhance local information for video lane segmentation. However, this method introduces attention to the entire image, making it memory-intensive and inefficient for autonomous driving scenarios.

\subsection{Homography Estimation}
Homography estimation methods can be divided into non-deep and deep learning-based approaches. Non-deep methods estimate the homography using feature extraction, feature matching, and outlier rejection. SIFT \cite{lowe2004distinctive}, SURF \cite{bay2006surf,bay2008speeded}, ORB \cite{rublee2011orb}, LPPM \cite{ma2019locality}, GMS \cite{bian2017gms}, and BEBLID \cite{suarez2020beblid} are commonly used for feature extraction, and RANSAC \cite{fischler1981random}, MAGAC \cite{barath2019magsac}, and LRLS \cite{holland1977robust} are applied for outlier rejection. Recently, many deep learning-based approaches have been proposed, such as DeTone \textit{et al.} \cite{detone2016deep}'s VGG-like architecture, Nowruzi \textit{et al.} \cite{erlik2017homography}, Le \textit{et al.} \cite{le2020deep}'s and Man \textit{et al.} \cite{man2019groundnet}'s cascaded VGG-like networks, and Chang \textit{et al.} and Zhao \textit{et al.}'s incorporation of the Lucas-Kanade (LK) algorithm with deep networks. However, these methods do not utilize known camera intrinsic and extrinsic parameters \footnote{In our experiments, both camera intrinsic and extrinsic parameters are provided by the dataset. In real-world scenarios, these parameters can be obtained through camera calibration and autonomous vehicle pose obtained from the pose estimation framework, which is already known in the in-car system and free of cost.} across frames for homography matrix estimation, which can effectively reduce the 
search dimensions.

\section{Method}
\subsection{Overview}
As illustrated in Fig.~\ref{fig:architecture}, our proposed framework takes a sequence of frames, denoted as $\mathbf{I} = \{\mathbf{I}_{i} | i = \{t - (s - 1) \Delta t\}_{s=1}^{n}\}$, as input.
This sequence consists of a current frame $\mathbf{I}_{t}$ and $n - 1$ previous frames sampled at a fixed time interval $\Delta t$. The output of the framework is a lane mark segmentation map $\mathbf{S}_{t}$ for the current frame. 
In addition, the framework includes a novel HomoFusion module, which uses the homography transformation guided by the proposed RSNE to fuse the feature map of the current frame with those of the previous frames. 
We denote the encoded feature representation for the $t^{th}$ frame as $\mathbf{F}_{t} = \{ \mathbf{F}_{t}^{l} \in \mathbb{R}^{H_{l}\times W_{l}\times C_{l}} \}_{l=1}^{L}$ where $l$ indicates the level of the feature map. The shallowest and deepest levels are represented by $1$ and $L$, respectively. $H_{l}$, $W_{l}$, and $C_{l}$ represent the height, width, and channel number of feature maps in  level $l$. We denote all encoded features as $\mathbf{F} = \{\mathbf{F}_{i}^{l}| i = \{t - (s - 1) \Delta t\}_{s=1}^{n}\}$.
The proposed HomoFusion module, detailed in Sec.~\ref{sec:homof}, 
uses a cross-frame pixel-to-pixel attention mechanism to fuse the feature map ($\mathbf{F}_{f} = \{ \mathbf{F}_{f}^{l} \}_{l=1}^{L}$) of the current frame with those of the previous frames. 
The homography transformation matrix between the target frame and each of the previous frames is calculated using the estimated normal vector of the road surface $\mathbf{n}$ obtained by the proposed RSNE, as described in Sec.~\ref{sec:normal}. Finally, the decoder produces the segmentation prediction ($\mathbf{S}_{t}$) for the target frame based on the fused feature map.

\subsection{Homography Guided Fusion (HomoFusion)}
\label{sec:homof}
\begin{figure}[t]
  \centering
  \includegraphics[width=0.99\linewidth]{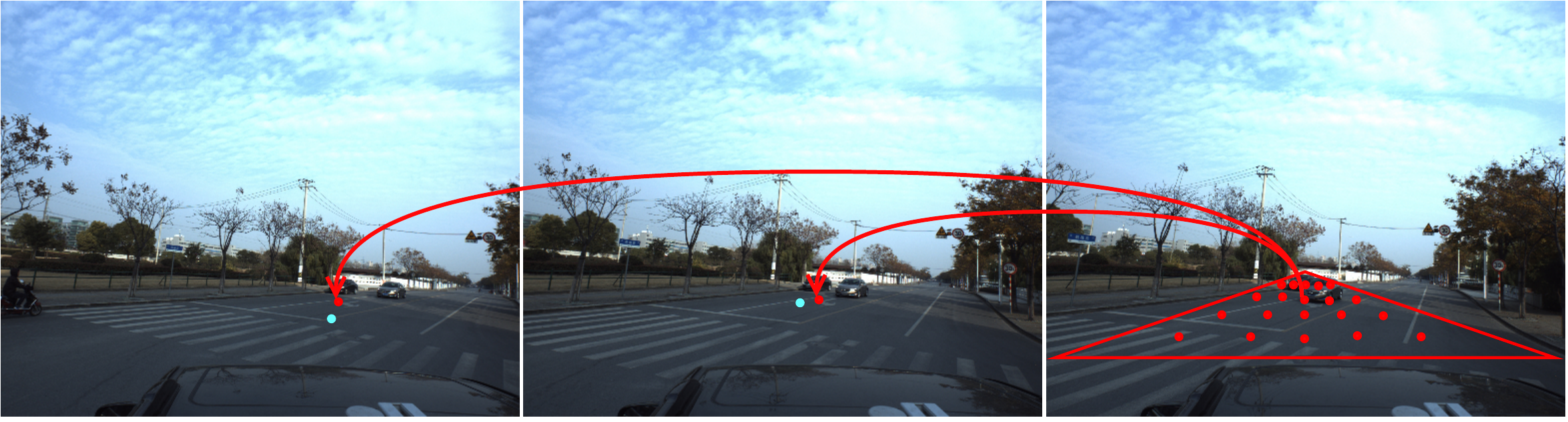}
  \caption{Illustration of sample points. (Right) \textcolor{red}{Sample points} in the current frame. (Left/Middle) Correspondence of a sample point in previous frames. The \textcolor{red}{red} point coordinate is calculated by using the correct normal, while the \textcolor{cyan}{cyan} point coordinate is calculated by using the initial (incorrect) normal.}
  \label{fig:point_sample}
    \vspace{-0.5cm}
\end{figure}

Our proposed HomoFusion module employs a pixel-to-pixel attention mechanism to fuse spatially corresponding pixels across frames. This mechanism is achieved by projecting the pixels of the current frame onto the previous frames through an accurate homography transformation. Since road lines and markings are strictly on-road, the search area can be limited to the road surface, which is mostly a plane, at least within the immediate front of the vehicle where markings are readable. Therefore, only the homography transformation is needed to accurately map road pixels between frames.  
According to the standard inverse homography \cite{Hartley2004,zhou2018stereo,tucker2020single}, for each on-road pixel $p_{t} = (u, v)^\top$ of the current frame, its spatially corresponding pixels of the reference frames $\{p_{i}| i = \{t - (s - 1) \Delta t\}_{s=1}^{n}\}$ can be computed using Eq.~\eqref{equ:homo}.

\begin{equation}
  \vspace{-0.2cm}
    p_{i} \propto \mathbf{K}(\mathbf{R}_{i} - \frac{\mathbf{t}_{i}\mathbf{n}^\top}{d} )\mathbf{K}^{-1}(p_{t} \oplus 1),
    \label{equ:homo}
\end{equation}
where $\propto$ represents proportional, $\mathbf{K}$ is the intrinsic matrix of the on-vehicle camera, $\mathbf{R}_{i}$ and $\mathbf{t}_{i}$ represent the relative rotation and translation between the current frame $\mathbf{I}_{t}$ and the $i^{th}$ previous frame $\mathbf{I}_{i}$. $\mathbf{n}$ is the normal of the observed road surface, which is estimated in Sec.~\ref{sec:normal}. $d$ is the vertical distance between the on-vehicle camera and the road surface obtained from the camera calibration. $\oplus 1$ converts the point coordinates to homogeneous coordinates. The on-road pixels of the current frame are sampled from a triangular area in front of the car, as illustrated in Fig.~\ref{fig:point_sample}. 
Figure \ref{fig:point_sample} also demonstrates that the shape and category of a left-turn can be recovered by exploring the spatially corresponding area in the previous frames under the guidance of an accurate homography transformation.

With the spatial correspondence across frames, the proposed HomoFusion module computes a temporally consistent representation for each sampled pixel of the current frame by employing temporal attention mechanism on the feature presentations of spatially corresponding pixels.  The mechanism assigns the pixel representation of current frame ($\mathbf{F}_{t}[p_{t}]$) as a query and all frames ($\{ \mathbf{F}_{i}[p_{i}]\}$) as keys and values. 
In contrast to the existing attention mechanism \cite{vaswani2017attention}, we propose to (1) omit the spatial encoding in the presence of obtained spatial correspondence across frames, and (2) apply an L2 normalization on the query and the keys to obtain more robust (deep) feature similarities (Eq.~\ref{equ:similarity}) across various lighting environments. 

\begin{equation}
    \mathbf{a}_{i} = \frac{\mathbf{F}_{t}[p_{t}]}{\|\mathbf{F}_{t}[p_{t}]\|_{2}}\cdot\frac{\mathbf{F}_{i}[p_{i}]}{\|\mathbf{F}_{i}[p_{i}]\|_{2}},
    \label{equ:similarity}
\end{equation}
where $\|.\|_{2}$ indicates an $L_{2}$ normalization. The similarity is normalized by softmax function as given in Eq.~\ref{equ:weight}: 
\begin{equation}
    \mathbf{W}_{i} = \frac{\exp{\mathbf{a}_{i}}}{\sum_i\exp{\mathbf{a}_{i}}},
    \label{equ:weight}
\end{equation}
before being used as weight to fuse the spatially corresponding pixels across the frames as shown in Eq.~\ref{equ:fusion}:
\begin{equation}
    \mathbf{F}_{f}[p_{t}] = \mathbf{F}_{t}[p_{t}] + \sum_{i}\mathbf{W}_{i}\mathbf{F}_{i}[p_{i}].
    \label{equ:fusion}
\end{equation}

\subsection{Road Surface Normal Estimator (RSNE)}
\label{sec:normal}
Accurate estimation of road surface normal is crucial for establishing spatial correspondence across frames. Generally, on-vehicle cameras have a fixed angle with respect to the ground level. However, 
the actual road surface normal can vary due to the unevenness of the road surface, \eg uphill or downhill roads, and sloped turns, \etc. 
Assuming that the initial road surface normal is a vector being perpendicular to a horizontal plane and pointing upwards,
we propose to obtain an accurate road surface normal with an 
optimization process that repeats for each new coming frame. 
The proposed RSNE iteratively updates the estimated road surface normal with the projection error of on-road pixels from the current frame onto the previous frames. Specifically, we sample $m$ pixels $\{p_{t}^{j}\}_{j=1}^{m}$ from the current frame that are likely to be on-road, as illustrated in Fig.~\ref{fig:point_sample}, and compute the spatially corresponding pixels $\{p_{i}^{j}\}_{j=1}^{m}$ in the previous frames using the homography matrix computed with the estimated road surface normal as defined in Eq.~\ref{equ:homo}. 
The estimated road surface normal is optimized in an iterative manner through the following procedure:
(a) computing a residual between the low-level features of spatially corresponding pixels across the frames with Eq.~\ref{equ:residual} 
\begin{equation}
    \mathbf{r}_{i}^{j} = \mathbf{F}_{i}^{1}[p_{i}^{j}] - \mathbf{F}_{t}^{1}[p_{t}^{j}] \in \mathbb{R}^{c},
    \label{equ:residual}
\end{equation}
where the residual is set to 0 for the sample point whose projections in previous frames fall outside the frame area, and computing the overall error with Eq.~\ref{equ:error}
\begin{equation}
    \mathbf{E} = \sum_{i,j}\rho(\lVert \mathbf{r}_{i}^{j} \rVert_{2}^{2}),
    \label{equ:error}
      \vspace{-0.2cm}
\end{equation}
where  $\|.\|_{2}^{2}$ shows squared norm and $\rho()$ is a robust cost function proposed by Barron \cite{Barron_2019_CVPR}. This nonlinear least-square cost is iteratively minimized towards correct normal estimation using the Levenberg–Marquardt algorithm \cite{levenberg1944method,marquardt1963algorithm}; (b) computing a new road surface normal value by adding the update 
given by Eq.~\ref{equ:delta}. 

The road surface normal vector is a unit vector with two degrees of freedom (DoF), which can be decomposed into a pitch angle $\theta$ and a roll angle $\phi$ \footnote{ An explanation of the pitch angle $\theta$ and roll angle $\phi$ is provided in the appendix.}, as shown in Eq.~\ref{equ:normal_angle}.
\begin{equation}
    \mathbf{n} = (-\sin{\phi}\cos{\theta}, ~-\cos{\phi}\cos{\theta}, ~\sin{\theta})^\top.
    \label{equ:normal_angle}
\end{equation}
The Jacobian of the residual function w.r.t. the pitch angle and the roll angle is defined in Eq.~\ref{equ:Jacobian} by applying the chain rule,
\begin{equation}
\mathbf{J}_i^{j} = \frac{\partial \mathbf{r}_{i}^{j}}{\partial\delta}=\frac{\partial \mathbf{F}_{i}^{1}[p_{i}^{j}]}{\partial p_{i}^{j}}\frac{\partial p_{i}^{j}}{\partial \mathbf{n}}\frac{\partial \mathbf{n}}{\partial\delta},
\label{equ:Jacobian}
\end{equation}
where $\delta$ represents the to-be-optimized target ($\theta$ or $\phi$), and $\frac{\partial \mathbf{F}_{i}^{1}[p_{i}^{j}]}{\partial p_{i}^{j}}$ is gradient from 2D interpolation. $\frac{\partial p_{i}^{j}}{\partial \mathbf{n}}$ is the Jacobian of the homography transformation (Eq.~\eqref{equ:homo}) \wrt $\mathbf{n}$, as defined in Eq.~\ref{equ:Jacobian_n}
\begin{equation}
\frac{\partial p_{i}^{j}}{\partial \mathbf{n}} = -\frac{1}{d} \mathbf{K}\mathbf{t}_{i}(\mathbf{K}^{-1}(p_{t}^{j} \oplus 1))^{\top},
\label{equ:Jacobian_n}
\end{equation}
and $\frac{\partial \mathbf{n}}{\partial\delta}$, defined in Eq.~\ref{equ:Jacobian_theta} and Eq.~\ref{equ:Jacobian_phi}, is the Jacobian of Eq.~\eqref{equ:normal_angle} \wrt $\theta$ and $\phi$ respectively
\begin{equation}
\frac{\partial \mathbf{n}}{\partial\theta} = (
-\frac{\mathbf{n}_{1}\mathbf{n}_{3}}{\sqrt{1-\mathbf{n}_{3}^{2}}},~ -\frac{\mathbf{n}_{2}\mathbf{n}_{3}}{\sqrt{1-\mathbf{n}_{3}^{2}}},~ \sqrt{1-\mathbf{n}_{3}^{2}} ),
\label{equ:Jacobian_theta}
\end{equation}
\begin{equation}
\frac{\partial \mathbf{n}}{\partial\phi} = ( \mathbf{n}_{2},~ -\mathbf{n}_{1},~ 0),
\label{equ:Jacobian_phi}
\end{equation}
where $\mathbf{n}_{k}$ represents the $k$-th element of $\mathbf{n}$. 
The detailed derivation is included in appendix (Sec.~1).
We obtain Hessian matrices as $\mathbb{H} = \mathbf{J}^\top\rho'\mathbf{J}$, where $\rho'$ is the derivative of function $\rho()$. We compute the update by damping the Hessian and solving the linear system, as shown in
\begin{equation}
\bigtriangleup\delta = -(\mathbb{H} + \lambda\text{ diag} (\mathbb{H}))^{-1}\mathbf{J}^\top\rho'\mathbf{r},
\label{equ:delta}
\end{equation}
where $\lambda$ is the damping factor \cite{sarlin21pixloc,wang2023satellite,Wang_2023_ICCV}, balancing between the Gauss-Newton ($\lambda=0$) and gradient descent ($\lambda=\infty$). Our entire optimization process is differentiable.

The algorithm is illustrated in Alg.~\ref{alg:normal_opt}. 
The normal vector $\mathbf{n}$ 
uses low-level features $\mathbf{F}^{1}$. 
While the high-level features $\mathbf{F}^{2}$ focus on segmentation feature expression, the low-level features $\mathbf{F}^{1}$ are supervised to extract useful features for both segmentation and road surface normal estimation. As both tasks focus on on-road objects, they can benefit each other.

\begin{algorithm}[t]
\caption{Normal Optimization}
\label{alg:normal_opt}
\textbf{Input:} \\
The $1$-st Level Feature Maps: $\mathbf{F}^{1} = \{\mathbf{F}_{i}^{1}\}$; \\
Sample Points Coordinates: $p_{t} = \{p_{t}^{j}\}_{j=1}^{m}$; \\
Initial Pitch and Roll Value: $(\theta_{0},~ \phi_{0}) = (0.15,~ 0)$;\\
Damping Factor: $\lambda = (\lambda_{\theta}~ \lambda_{\phi})$; \\
Hyper-parameter: \\
$\mathbf{k}$; \Comment{Maximum loop count, empirically set to 20} \\ 
$\alpha$; \Comment{Convergence threshold, empirically set to 0.0001} \\
\textbf{Output:} 
Optimized Pitch and Roll $(\theta,~ \phi)$; \\
\begin{algorithmic}[1]
    \Function{OPT}{$\theta_{0}, \phi_{0}, \mathbf{F}^{1}, p_{t}, \lambda$}
    \State Derive point features $\mathbf{F}_{t}^{1}[p_{t}]$ from $\mathbf{F}_{t}^{1}$;
    \For{$k \gets 1$ to $\mathbf{k}$}
        \State Calculate point coordinates $p_{i}$ in $\mathbf{F}_{i}^{1}$ (Eq.~\ref{equ:homo});
        \State Derive point features $\mathbf{F}_{i}^{1}[p_{i}]$ from $\mathbf{F}_{i}^{1}$;
        \State Calculate residual $\mathbf{r}_{i}$ (Eq.~\ref{equ:residual});
        \State Calculate observe error $\mathbf{E}$  (Eq.~\ref{equ:error});
        \State Calculate robust cost $\rho(\mathbf{E})$ and its derivation $\rho'$;
        \State Construct $\mathbf{J}$ and its Hessian matrices $\mathbb{H}$ (Eq.~\ref{equ:Jacobian});
        \State Obtain $\bigtriangleup\delta$ by Cholesky decomposition (Eq.~\ref{equ:delta});
        \State Update Normal as $(\theta,~ \phi)_{k}  \gets \ (\theta,~ \phi)_{k-1}+ \bigtriangleup\delta$;
        \If{MAX$(\bigtriangleup\delta)<\alpha$}
            \State Break;
        \EndIf
    \EndFor
    \EndFunction
\end{algorithmic}
\end{algorithm}

\begin{table*}[t] \small
\renewcommand{\arraystretch}{0.90}
\caption{Comparison with SOTA Methods on the ApolloScape
\cite{huang2018apolloscape} and Apolloscape Night Datasets and Running on an NVIDIA RTX 3090 GPU. ``18 mIoU" Represents the Mean IoU of 18 Types of Lane Markers Selected by ApolloScape Official Metrics. ``36 mIoU" Represents the Mean IoU of All Unignorable 36 Types. }
\vspace{-0.2cm}
\label{Tab:Comparison}
\centering
\setlength\tabcolsep{1pt}
\begin{tabular*}{\textwidth}{c|c|c@{\extracolsep{\fill}}|c|c|c| c c |c c }
\hline
\multirow{2}{*}{Methods}& \multirow{2}{*}{Frame Count} & \multirow{2}{*}{Backbone} & 
\multirow{2}{*}{Params (M)$\downarrow$} &  
\multirow{2}{*}{GFLOPs$\downarrow$}& 
\multirow{2}{*}{FPS (f/s)$\uparrow$} &
\multicolumn{2}{c|}{ApolloScape \cite{huang2018apolloscape}} & \multicolumn{2}{c}{ApolloScape Night} \\
& & & & & &18 mIoU$\uparrow$ & 36 mIoU$\uparrow$ & 18 mIoU$\uparrow$ & 36 mIoU$\uparrow$ \\
\hline
IntRA-KD\cite{hou2020inter} & 1 & ResNet-101 & 65.6 & 5159.4 & 10.8 & 42.1 & 24.6 & 29.8 & 16.7 \\
SegFormer\cite{xie2021segformer}& 1 & MiT-B1 & 13.5 & 1048.8 & \textbf{43.8}  & 52.3 & 32.1 & 38.3 & 23.1 \\
CFFM \cite{sun2022vss} & 4 & MiT-B1 & 15.3 & 1192.6 & 22.7 & 53.2 & 32.7 & 39.1 & 23.6\\
MMA-Net \cite{zhang2021vil} & 4 & ResNet-50 & 57.9 & 723.2 & 20.6 & 52.9 & 31.4 & 38.8 & 23.2 \\
HomoFusion(ours) & 4 & EfficientNet-B6 & \textbf{1.24} & \textbf{61.2} & 25.4 & \textbf{59.3} & \textbf{35.9} & \textbf{44.8} & \textbf{26.6}\\
\hline
\end{tabular*}
\vspace{-0.5cm}
\end{table*}

\section{Experiments}
\subsection{Implementation Details}
\label{sec:implementation}
We use a pre-trained EfficientNet \cite{tan2019efficientnet} to extract image features at two different scales, $4\times$ and $16\times$ down-scaling, with $64$ and $128$ channels, respectively. To focus on the bottom region of the input image, where road marks are typically visible, we crop only the bottom $40\%$ of the input image, following the same approach in the prior work \cite{hou2020inter}. 
The processed image size is set to $272 \times 848$. 
Our decoder consists of bi-linear up-sampling layers and convolution layers, which up-sample the high-level features and increase the resolution by a factor of $4$ at each level.
We trained our model using the AdamW optimizer \cite{you2019large} for 30,000 iterations on two NVIDIA RTX 3090 GPUs, with a learning rate of $4 \times 10^{-3}$.

\subsection{Datasets}
\vspace{-0.2cm}
We conduct our experiments on two datasets, ApolloScape \cite{huang2018apolloscape} and ApolloScape Night.
Other datasets were deemed unsuitable due to the absence of camera extrinsics, as observed in SDLane \cite{Jin2022eigenlanes} and VLI-100 \cite{zhang2021vil}, inadequate frame overlap, or the lack of diverse segmentation labels, exemplified by Waymo Open Dataset \cite{sun2020scalability}. Further elaboration on these reasons can be found in the appendix.

\noindent\textbf{ApolloScape}. The ApolloScape dataset is a large-scale dataset that can be used for localization and segmentation tasks. It consists of $38$ distinct classes and poses various challenges including occlusions and tiny road markings. 
To provide accurate camera poses, the vehicle is equipped with customer-grade GPS/IMU. However, for the lane mark segmentation task, only $41,201$ annotated images out of more than $110,000+$ are associated with camera pose information. We used these $41,201$ images for our experiments since our approach relies on camera poses. We divided them into $35,173$ training images and $6,028$ validation images. 
Because our approach uses sequential information, we select the validation set with completely different trajectories to ensure a fair comparison. 

\begin{figure}[t]
  \centering
  \includegraphics[width=0.99\linewidth]{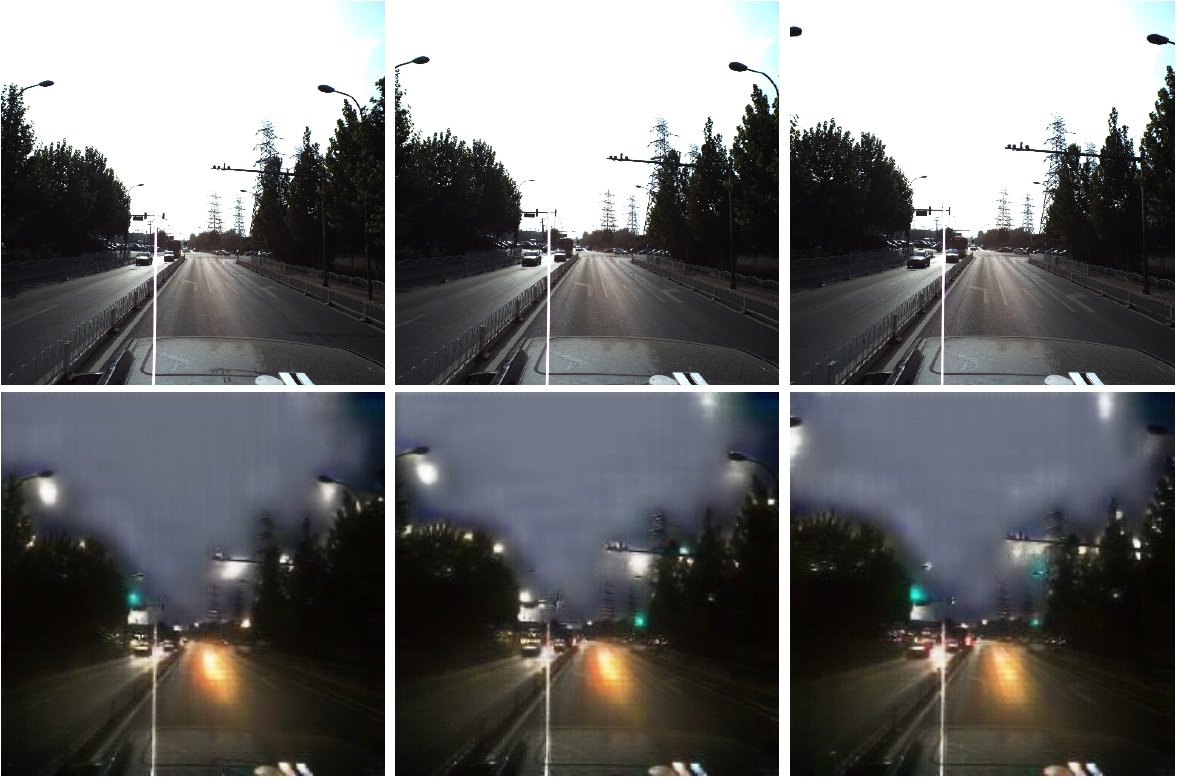}
  \caption{Sample images from the ApolloScape Night dataset. Top: original daytime images from the ApolloScape dataset. Bottom: synthesized night-time images.}
  \label{fig:night_dataset}
\vspace{-15pt}
\end{figure}

\noindent\textbf{ApolloScape Night}.
Since there is a lack of datasets for night-time lane mark segmentations, we created an artificial dataset called ApolloScape Night from the ApolloScape dataset using a cross-domain generation network \cite{arruda2019cross}. 
This allows us to evaluate our proposed model on a challenging dataset with poor lighting conditions, road reflection, and glare.
Fig.~\ref{fig:night_dataset} displays some sample images from the dataset.

\subsection{Evaluation Metrics}
In accordance with the guidelines of the ApolloScape benchmark \cite{huang2018apolloscape}, we used mean intersection-over-union (mIoU) as the evaluation criterion. ApolloScape contains 38 different types of lane markers, including two ignorable labels (noise and ignored). Of these labels, 18 categories are used in the official evaluation metrics.
To provide a comprehensive evaluation of our approach, we report the evaluated mIoU results for both the selected 18 categories, denoted as \enquote{18 mIoU}, and all 36 categories, denoted as\enquote{36 mIoU}. 
By reporting both sets of results, we provide a more complete picture of the performance of our method on the ApolloScape and ApolloScape Night datasets.

\subsection{Comparison with State-of-the-art Methods}

\begin{figure*}[!htb]
    \centering
    \includegraphics[width=0.99\textwidth]{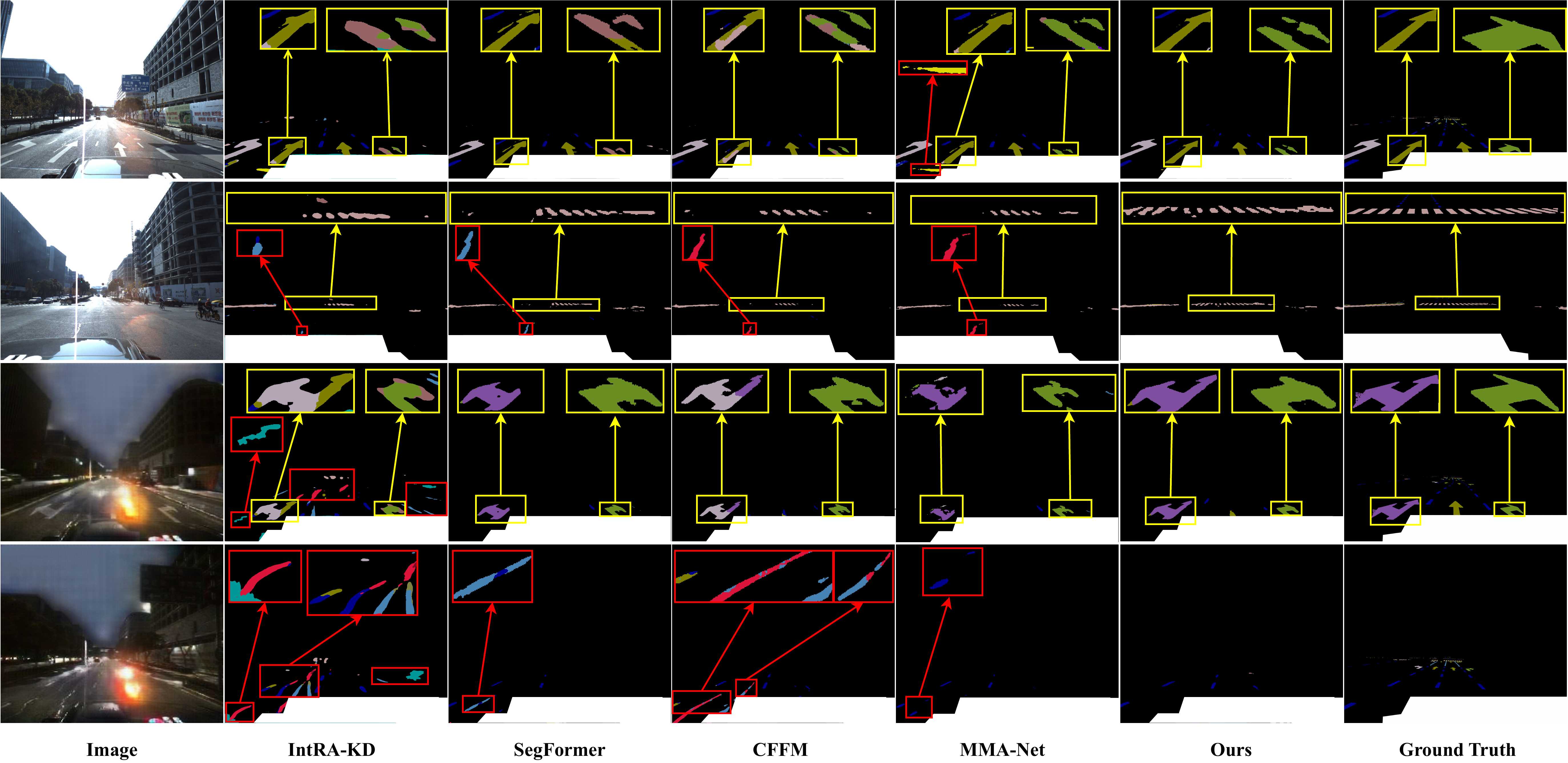}
    \caption{Qualitative comparison with SOTA methods. The top two examples are from the ApolloScape \cite{huang2018apolloscape} dataset, and the bottom two examples are from the ApolloScape Night dataset. Yellow boxes highlight the area of interest for better visualization. Red boxes indicate false-positive segmentation predictions. Best viewed in color.}
    \label{fig:compare_vis}
    \vspace{-0.5cm}
\end{figure*}

To evaluate the performance of our proposed method, we compared it with SOTA algorithms, including (a) IntRA-KD \cite{hou2020inter},  (b) SegFormer \cite{xie2021segformer}, (c) CFFM \cite{sun2022vss}, and (d) MMA-Net \cite{zhang2021vil}, on the ApolloScape \cite{huang2018apolloscape} and ApolloScape Night datasets.
To ensure a fair comparison, we retrained each model on the same subset of the training set with the same input resolution.

The results of the comparison are presented in Table \ref{Tab:Comparison}. 
Our approach outperforms the existing SOTA methods on both the ApolloScape \cite{huang2018apolloscape} and ApolloScape Night datasets while having less than 9\% of their parameters or computational overhead.

In addition, we compared our proposed method with CFFM \cite{sun2022vss} which also uses adjacent frames to enhance the representation of the current frame. 
Our method is more efficient in terms of model complexity, with a complexity of  $\mathcal{O}(HWC)$ for feature extraction and $\mathcal{O}(HWC^2)$ for cross-frame feature fusion.
In contrast, the complexity of CFFM is  $\mathcal{O}(H^2W^2C)+\mathcal{O}(HWC^2)$ for feature extraction and $\mathcal{O}(HWEC)+\mathcal{O}(HWC^2)$ for cross-frame feature fusion, where $E$ is calculated by their receptive field and pooling kernel size. 
This explains why our method has a lower computational requirement for the cross-frame pixel-to-pixel attention mechanism compared to the global attention mechanism used by CFFM.

\begin{figure}[t]
    \centering
    \subfloat[Number of frames (AS))]{
        \includegraphics[width=0.21\textwidth]{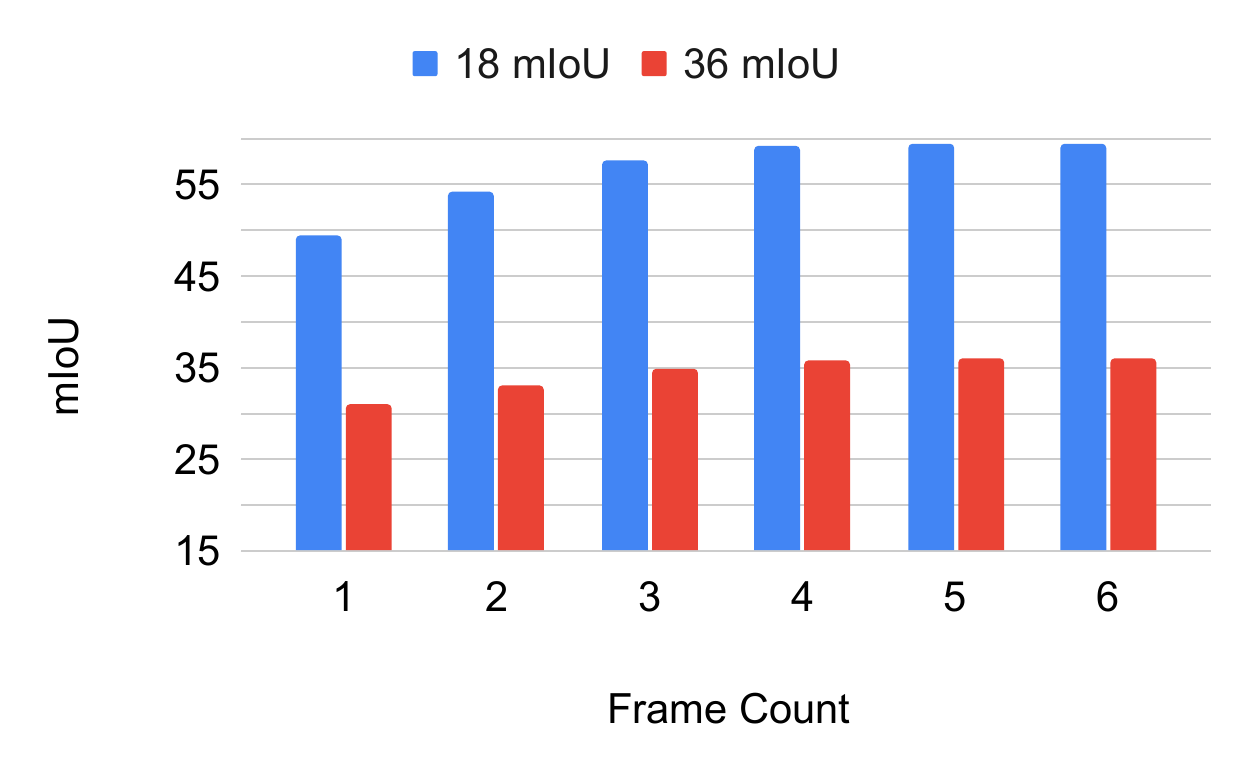}
    }
    \subfloat[Number of frames (ASN)]{
        \includegraphics[width=0.21\textwidth]{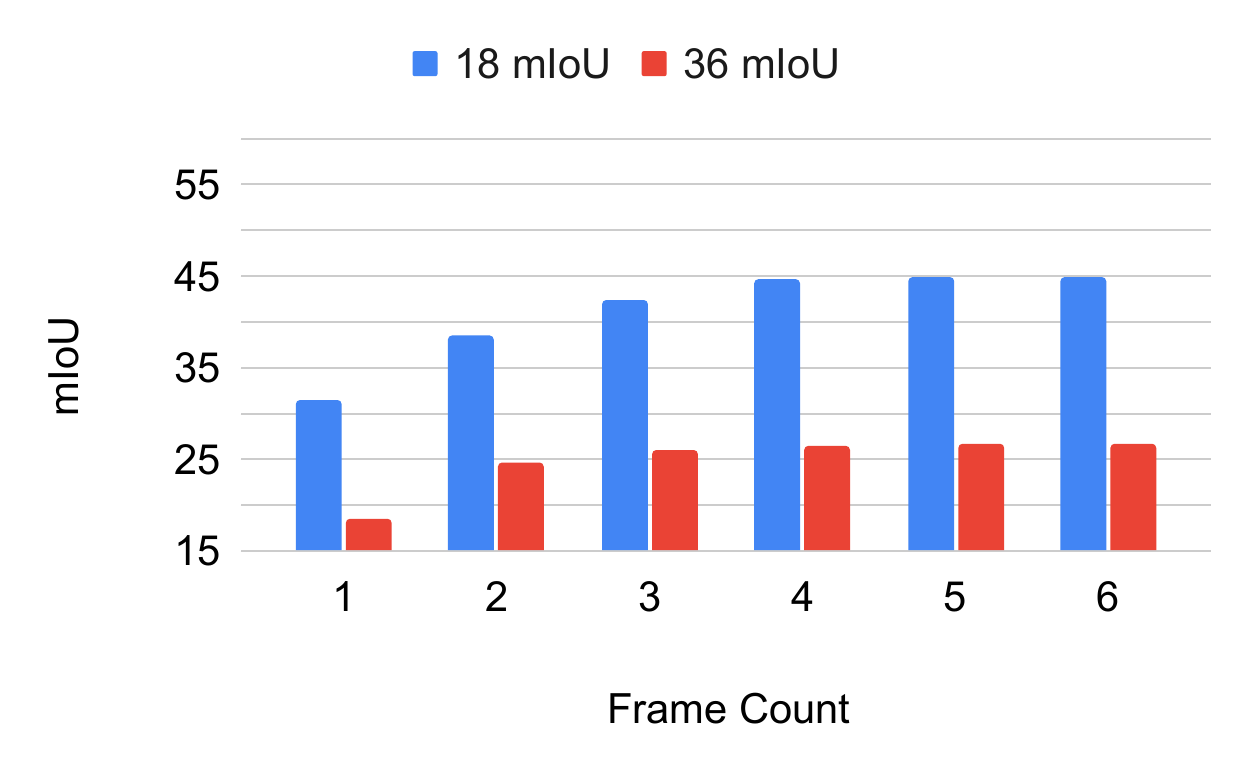}
    }
    \\
    \subfloat[Sampling frame gap (AS)]{
        \includegraphics[width=0.21\textwidth]{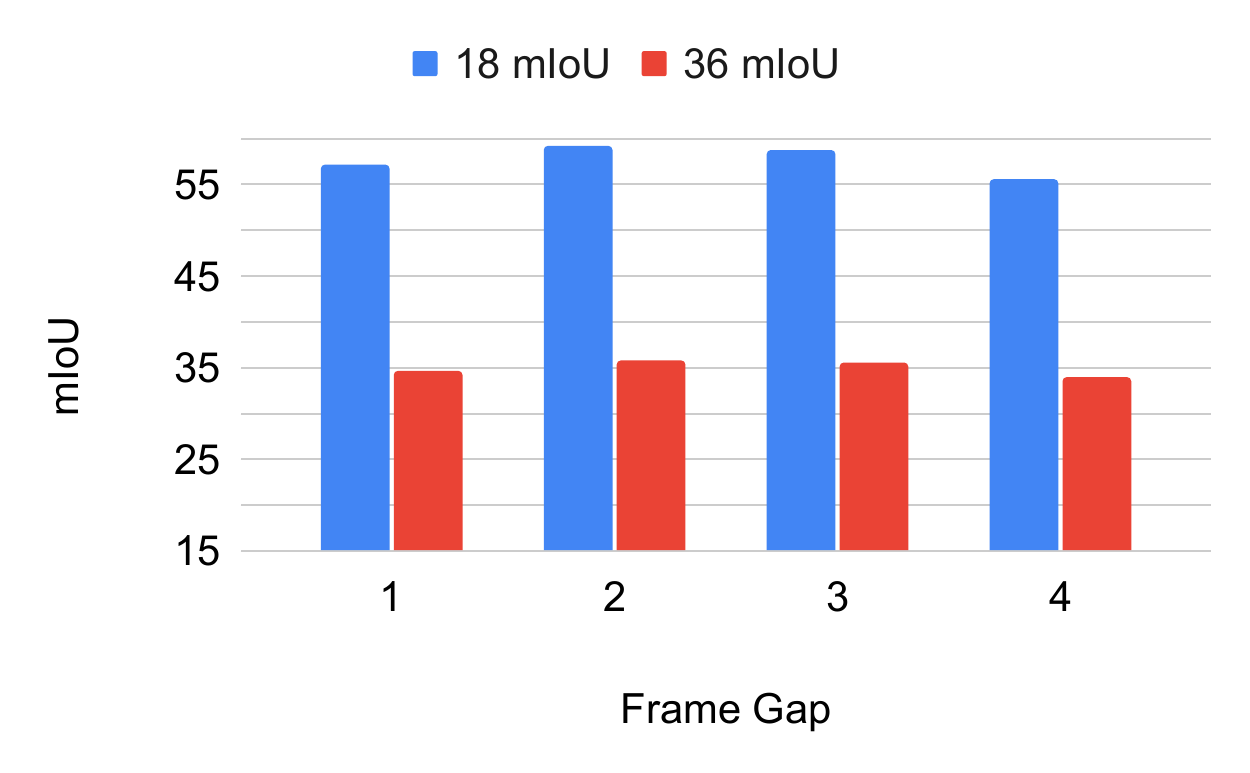}
    }
    \subfloat[Sampling frame gap (ASN)]{
        \includegraphics[width=0.21\textwidth]{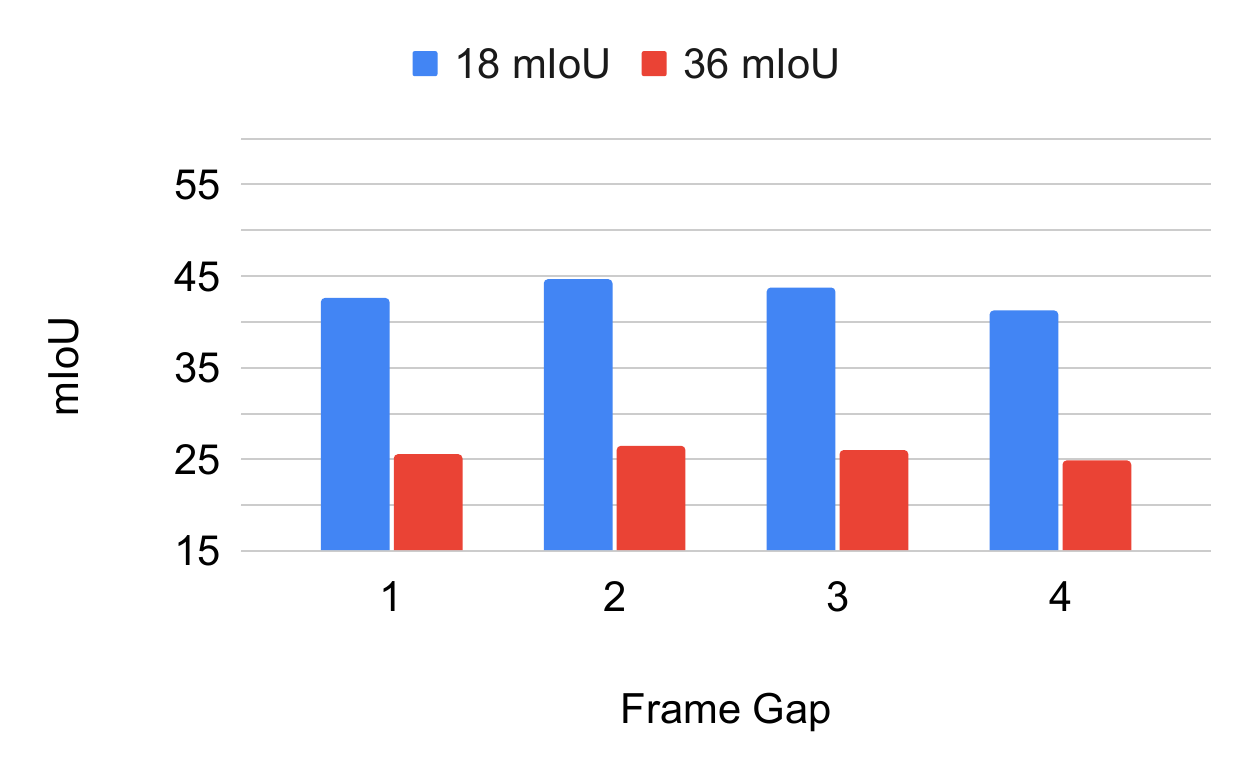}
    }
    \caption{
    The impact of the number of frames $n$ on ApolloScape (AS) and ApolloScape Night (ASN) datasets is shown in (a) and (b), while (c) and (d) show the effect of the sampling frame gap $\bigtriangleup t$.
    }
    \label{fig:hyperpara}
    \vspace{-0.6cm}
\end{figure}

The qualitative results of both the existing SOTA models and our proposed model are shown in Fig.~\ref{fig:compare_vis}.
The results indicate that our approach achieves superior segmentation performance on road lines and markings, even under adverse conditions such as occlusion, road reflection, and poor lighting. Additional visualization examples for various categories can be found in the appendix.

\subsection{Study on Hyper-parameters }
\noindent\textbf{Number of Frames ($n$)}. 
We conducted a study on the impact of the number of frames while setting the sampling frame gap $\bigtriangleup t = 2$.
The results presented in Fig.~\ref{fig:hyperpara} (a)-(b) demonstrate that the performance of our method improves on both ApolloScape and ApolloScape Night datasets with an increase in the number of explored adjacent frames. However, the performance improvement becomes negligible when the number of frames exceeds $4$. Considering the monotonically increasing relationship between the model complexity and the number of frames, we set $n = 4$ to strike a balance between model performance and complexity.

\noindent\textbf{Sampling Frame Gap ($\bigtriangleup t$)}. In the study on the effect of the sampling frame gap, we set the number of frames $n = 4$. Fig.~\ref{fig:hyperpara} (c)-(d) indicates that the optimal performance is achieved at $\bigtriangleup t = 2$.
This can be attributed to the balance between the sufficient overlap of road area across the frames and a wide enough temporal gap that allows the displacement of the source of occlusion, \eg moving vehicle.

\subsection{Discussion}
All ablation studies are conducted on the ApolloScape dataset with the same training strategies described in Sec.~\ref{sec:implementation}. We set the number of frames to $n = 4$ and a sampling frame gap of $\bigtriangleup t = 2$.

\noindent\textbf{Impact of HomoFusion}.
The core component of our approach is the HomoFusion module, which allows the use of complementary information from sequential frames. Tab.~\ref{Tab:Ablation_study} demonstrates that our model with the HomoFusion module (4 frames) achieves better performance than the variant without HomoFusion module (1 frame), illustrating the effectiveness of the temporally consistent feature representation obtained with HomoFusion module in addressing the partial occlusion issues. We also study the effect directly associating the pixels across frames based on their coordinates 
by using identity matrix as the homography matrix, denoted as \enquote{w/o HomoGuide}. It shows that such an incorrect cross-frame spatial correspondence, although leveraging additional frames, provides limited help in improving the model performance.

\begin{table}[t]
\caption{Ablation Study of our Proposed HomoFusion and RSNE on the ApolloScape Dataset
}
\vspace{-0.1cm}
\label{Tab:Ablation_study}
\centering
\footnotesize
\renewcommand{\arraystretch}{1.0}
\begin{tabular}{c|@{\extracolsep{\fill}}ccc}
\hline
& 18 mIoU$\uparrow$ & 36 mIoU$\uparrow$ \\
\hline
w/o HomoFusion & 49.4 & 31.0\\
HomoFusion w/o HomoGuide & 51.8 & 31.3  \\
HomoFusion w/o RSNE & 57.4 & 35.3 \\ 
HomoFusion w/ RGB RSNE & 54.9 & 33.8 \\ 
Noisy Extrinsic & 57.5 & 35.1 \\
HomoFusion Full & \textbf{59.3} & \textbf{35.9} \\
\hline
\end{tabular}
\vspace{-0.6cm}
\end{table}

\begin{figure}[h]
    \centering
     \vspace{-0.4cm}
    \subfloat[Normal estimation with RGB]{
        \includegraphics[width=0.24\textwidth]{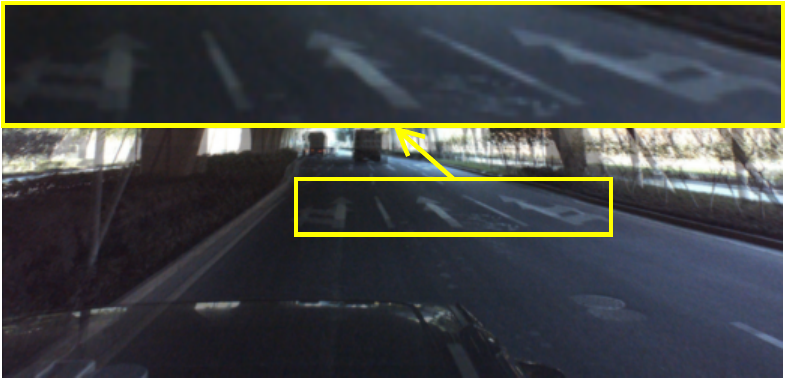}
    }
    \subfloat[Normal estimation with feature ]{
        \includegraphics[width=0.24\textwidth]{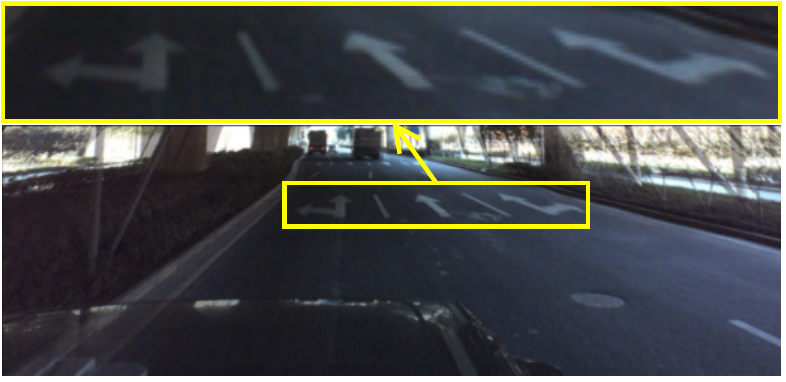}
    }
    \caption{\small Fused frame sequence with the homography transformation between the frame pairs. The road surface normal estimated with feature representation enables more accurate homography transformation than with RGB information, resulting in better alignment of road lines and markings across frames.
    }
    \vspace{-0.2cm}
    \label{fig:normal_impact}
\end{figure}

\noindent\textbf{Impact of RSNE}. We study the impact of the proposed RSNE in the following model variants each comprising the HomoFusion modules: (a) without RSNE (\enquote{HomoFusion w/o RSNE}), where the homography matrix is estimated with the initial road surface normal ($\mathbf{n}_{0}$), (b) with RSNE using RGB information (\enquote{HomoFusion w/ RGB RSNE}) and (c) with RSNE with features (\enquote{HomoFusion Full}).  
Results in Tab.~\ref{Tab:Ablation_study} show that \enquote{HomoFusion Full} achieves better performance than \enquote{HomoFusion w/o RSNE}, demonstrating the effectiveness of the iterative optimization in estimating a more accurate homography matrix that subsequently provides more accurate guidance on cross-frame spatial correspondences. It can be also observed that \enquote{HomoFusion Full} significantly outperforms \enquote{HomoFusion w/ RGB RSNE} indicating  that deep features are more robust than RGB information in estimating an accurate road surface normal. This can be attributed to the fact the feature representations are less prone to adverse noises caused by environmental factors, \eg brightness.
Fig.~\ref{fig:normal_impact} shows that the fused frame with road surface normal estimated with features has better alignment in road lines and markings than that with RGB information.
 
\noindent\textbf{Robustness against Extrinsic Noise}. 
We investigate the robustness of our method to extrinsic noise by introducing random translation errors of less than 1 meter and rotation errors of up to $30^\circ$. The results are shown in \enquote{Noisy Extrinsic} of Tab.~\ref{Tab:Ablation_study}, which demonstrate that our method is capable of mitigating extrinsic noise and remains robust.

\noindent\textbf{Impact of Backbone}. In order to ensure a fair comparison with CFFM \footnote{We use a heavy backbone in the competing methods as their publicly available code uses this backbone} \cite{sun2022vss} and MMA-Net \cite{zhang2021vil}, we implement our method with `MiT-B1' and 'ResNet-50' backbone and report its performance in Tab.~\ref{Tab:backbone}. Our method still  outperforms CFFM \cite{sun2022vss} and MMA-Net \cite{zhang2021vil} significantly. Additionally, we test our method with  
‘EfficientNet-B4/5’ backbones, and the results show that our method has relatively stable performance even when using smaller backbones.

 \begin{table}[h]
 \footnotesize 
\caption{Comparison of our Proposed HomoFusion with Different Backbones on ApolloScape Dataset
}
\vspace{-0.1cm}
\label{Tab:backbone}
\centering
\renewcommand{\arraystretch}{1.0}
\begin{tabular}{@{\extracolsep{\fill}}c|c|c|c|c}
\hline
Backbones & Params(M)$\downarrow$ & GFLOPs$\downarrow$ & 18 mIoU$\uparrow$ & 36 mIoU$\uparrow$ \\
\hline
MiT-B1 & 13.64 & 89.8 & 57.2 & 35.3 \\
ResNet-50 & 9.12 & 189.7 & 55.3 & 34.6\\
EfficientNet-B4 & \textbf{0.8} & \textbf{60.8} & 57.0 & 35.3\\
EfficientNet-B5 & 1.02 & 61.2 & 57.9 & 35.2\\
EfficientNet-B6 & 1.24 & 61.2 & \textbf{59.3} & \textbf{35.9}\\
\hline
\end{tabular}
\vspace{-0.5cm}
\end{table}

\noindent\textbf{Application to Another Task}.
Given our method's ability to accurately align road surface objects across adjacent frames, we also try it with the task of detecting water hazards. Specifically, our approach aims to perform binary segmentation to identify water puddles in the road. The previous work on this task T3D-FCN \cite{8945849} also exploits temporal information, as water puddles can appear differently at different angles and distances as a vehicle moves. 
We use the Puddle-1000 dataset \cite{han2018single}, which includes On-road and Off-road datasets. The frames obtained from an onboard camera on bumpy roads are not stable and contain small extrinsic noise. The intrinsic information of the camera is provided by the dataset, while the global extrinsic information of the camera is obtained using ORB-SLAM \cite{mur2015orb}.
Our experiments are conducted using the current and seven previously consecutive frames with a $240 \times 320$ image size, the same settings as T3D-FCN \cite{8945849}.
The results, presented in Tab.~\ref{Tab:water_hazard}, demonstrate the high effectiveness of our method.
More visual details of this task can be found in the Appendix.

\begin{table}[h]
\small
\caption{Comparison of water puddle segmentations on Puddle-1000 dataset
}
\label{Tab:water_hazard}
\centering
\footnotesize 
\renewcommand{\arraystretch}{1.}
\begin{tabular}{c|c|@{\extracolsep{\fill}}ccc}
\hline
 Dataset&Methods & F1-meas$\uparrow$ & Prec$\uparrow$ & Rec$\uparrow$ \\
\hline
\multirow{3}{*}{On-road} 
&FCN-8s-FL-RAU \cite{han2018single} &  0.70&  0.68& 0.72\\
&T3D-FCN \cite{8945849} &  0.68&  0.79& 0.62\\
&HomoFusion (ours) & \textbf{0.80} &  \textbf{0.81}& \textbf{0.78}\\
\hline
\multirow{3}{*}{Off-road} 
&FCN-8s-FL-RAU \cite{han2018single} &  0.81&  0.87& 0.77\\
&T3D-FCNU \cite{8945849} &  0.73&  0.87& 0.63\\
&HomoFusion (ours) &  \textbf{0.87}&  0.87& \textbf{0.87}\\
\hline
\end{tabular}
\vspace{-0.6cm}
\end{table}


\section{Conclusion}
The proposed lightweight lane mark segmentation model presented in this paper offers superior performance with reduced model complexity compared to SOTA approaches. The integration of the HomoFusion module and Road Surface Normal Estimator provides an accurate classification of partially occluded, shadowed, and/or glare-affected road lines and markings by leveraging adjacent frames and pixel-to-pixel attention mechanisms. Moreover, the novel approach of using the road surface normal to guide spatial correspondences across frames has significant implications for a wide range of applications in autonomous driving, including road surface inspection, water, and other hazard detections. However, the current model relies on available camera extrinsics and is somewhat impacted by extrinsic noise. Future work will focus on optimizing both the normal vector and camera extrinsic parameters to improve the performance and make the method more robust to challenging environmental conditions. Overall, the proposed method has the potential to significantly advance the field of autonomous driving and related applications.

\section{Acknowledgements}
The research is funded in part by an ARC Discovery Grant (grant ID: DP220100800) to HL.



{\small
\bibliographystyle{ieee_fullname}
\bibliography{egbib}
}

\clearpage
\appendix
\etocdepthtag.toc{mtappendix}
\etocsettagdepth{mtchapter}{none}
\etocsettagdepth{mtappendix}{subsection}
\tableofcontents
\clearpage

\input{supplementary}

\end{document}

%% file: supplementary.tex
\section{Derivation of Jacobian}
\begin{figure}[!h]
      \centering
      \subfloat[Pitch: rotation around x-axis]{
          \includegraphics[width=0.24
 \textwidth]{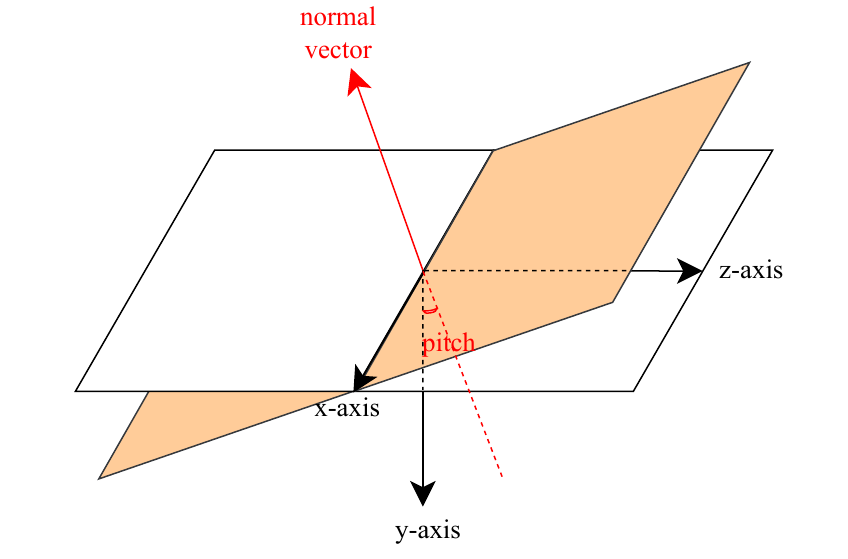}}
     \subfloat[Roll: rotation around  z-axis]{
           \includegraphics[width=0.24
 \textwidth]{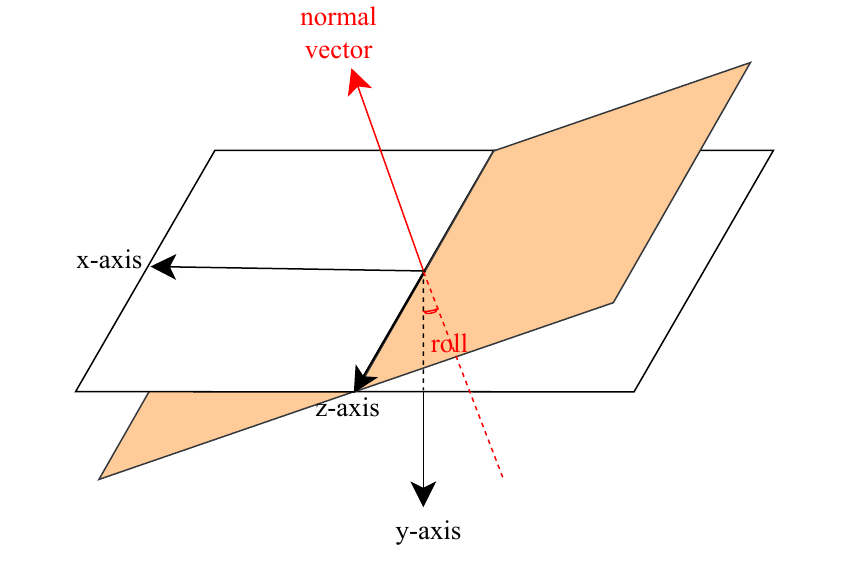}
  }
    \caption{Pitch and Roll of Road Surface Normal Vector Components}
    \label{fig:pitch_roll}
\end{figure}

In this section, we present the detailed derivations for the Equation.~9, Equation.~10 and Equation.~11 of the main paper, which calculate the Jacobian of Homography Transformation \wrt the pitch $\theta$ and the roll $\phi$ angle. 
The pitch angle $\theta$ and the roll angle $\phi$ represent the rotation of the road surface \wrt the plane constructed with the x-axis and z-axis of camera coordinates, as shown in Fig.~\ref{fig:pitch_roll}. 

\noindent\textbf{Derivation of Equation.~9 of the Main Paper} is the Jacobian of Homography Transformation (Equation.~1 of the Main Paper) \wrt road surface normal $\mathbf{n}$.
\begin{equation}
\begin{aligned}
\frac{\partial p_{i}^{j}}{\partial \mathbf{n}} &= \frac{\partial (\mathbf{K}(\mathbf{R}_{i} - \frac{\mathbf{t}_{i}\mathbf{n}^\top}{d} )\mathbf{K}^{-1}(p_{t} \oplus 1))}{\partial \mathbf{n}} 
\\
&= \frac{\partial (\mathbf{K}\mathbf{R}_{i}\mathbf{K}^{-1}(p_{t} \oplus 1))}{\partial \mathbf{n}} - \frac{\partial (\mathbf{K}( \frac{\mathbf{t}_{i}\mathbf{n}^\top}{d} )\mathbf{K}^{-1}(p_{t} \oplus 1))}{\partial \mathbf{n}} 
\\
&= 0 - \frac{1}{d} \frac{\partial (\mathbf{K}\mathbf{t}_{i}\mathbf{n}^\top\mathbf{K}^{-1}(p_{t} \oplus 1))}{\partial \mathbf{n}} 
\\
&= -\frac{1}{d} \mathbf{K}\mathbf{t}_{i}(\mathbf{K}^{-1}(p_{t}^{j} \oplus 1))^{\top},
\end{aligned}
\label{equ:Jacobian_n_detail}
\end{equation}

\noindent\textbf{Derivation of Equation.~10 of the Main Paper} is the Jacobian of road surface normal \wrt the pitch angle $\theta$. 

\begin{equation}
\begin{aligned}
\frac{\partial \mathbf{n}}{\partial\theta} &= (\frac{\partial(-\sin{\phi}\cos{\theta})}{\partial\theta}, \frac{\partial(-\cos{\phi}\cos{\theta})}{\partial\theta}, \frac{\partial(\sin{\theta)}}{\partial\theta})
\\
&= (\sin{\phi}\sin{\theta}, \cos{\phi}\sin{\theta}, \cos{\theta})
\\
&= ( \frac{\sin{\phi}\cos{\theta}\sin{\theta}}{\cos{\theta}}, \frac{\cos{\phi}\cos{\theta}\sin{\theta}}{\cos{\theta}}, \cos{\theta})
\\
&= ( -\frac{\mathbf{n}_{1}\mathbf{n}_{3}}{\cos{\theta}}, -\frac{\mathbf{n}_{2}\mathbf{n}_{3}}{\cos{\theta}}, \cos{\theta})
\\
&=(
-\frac{\mathbf{n}_{1}\mathbf{n}_{3}}{\sqrt{1-\mathbf{n}_{3}^{2}}},~ -\frac{\mathbf{n}_{2}\mathbf{n}_{3}}{\sqrt{1-\mathbf{n}_{3}^{2}}},~ \sqrt{1-\mathbf{n}_{3}^{2}} ),
\end{aligned}
\label{equ:Jacobian_theta_detail}
\end{equation}

\noindent\textbf{Derivation of Eq.~11 of the Main Paper} is the Jacobian of road surface normal \wrt the roll angle $\phi$. 

\begin{equation}
\begin{aligned}
\frac{\partial \mathbf{n}}{\partial\phi} &= 
(\frac{\partial(-\sin{\phi}\cos{\theta})}{\partial\phi}, \frac{\partial(-\cos{\phi}\cos{\theta})}{\partial\phi}, \frac{\partial(\sin{\theta)}}{\partial\phi})
\\
&= (-\cos{\phi}\cos{\theta},\sin{\phi}\cos{\theta}),0)
\\
&= ( \mathbf{n}_{2},~ -\mathbf{n}_{1},~ 0),
\end{aligned}
\label{equ:Jacobian_phi_detail}
\end{equation}

\section{Performance with Estimated Camera Extrinsic}
One limitation of our method is that it requires additional camera extrinsic information. Although it is easy to obtain in a real vehicle system, it may not be available for some datasets. To explore our performance under extrinsic unknown settings, we estimated the camera extrinsic using COLMAP \cite{schoenberger2016sfm} and used the estimated extrinsic to evaluate the proposed method. 
However, the ApolloScape dataset's larger camera elevation angle caused the key points obtained by COLMAP to be further away from the vehicle, resulting in noisy camera poses. Despite these less accurate poses, our method still achieved plausible performances ( \enquote{w/ COLMAP Extrinsic} in Tab.~\ref{Tab:noise}). 

\section{Performance with Estimated Homography}
We also use FindHomography function from OpenCV package to directly estimate the homography transformation matrix. The estimations are less accurate than our HomoGuide, leading to inferior lane segmentation results (\enquote{w/ OpenCV Homography} of Tab.~\ref{Tab:noise}).

\begin{table}[h]
\caption{Comparison of Estimated Extrinsic and OpenCV Homography on ApolloScape Dataset
}
\label{Tab:noise}
\centering
\renewcommand{\arraystretch}{1.}
\begin{tabular}{c|@{\extracolsep{\fill}}ccc}
\hline
 & 18 mIoU$\uparrow$ & 36 mIoU$\uparrow$ \\
\hline
HomoFusion (ours) & \textbf{59.3} & \textbf{35.9} \\
w/ COLMAP Extrinsic & 56.2 & 34.8 \\
w/ OpenCV Homography & 47.9 & 29.4 \\
\hline
\end{tabular}
\end{table}

\section{Challenging Scenarios}
In this section, we evaluate the performance of our method on challenging scenarios. As we utilize the LM algorithm for road surface normal vector estimation, there is a convex range of the algorithm that can result in failure due to excessive re-projection errors. To further investigate this scenario, we selected the top 100 inputs with the highest residuals after optimization. We compared our approach with and without RSNE and found that incorporating the RSNE led to improved performance (18 mIoU: 53.3 and 52.5, respectively), even in cases where the optimization has a high possibility of being non-convex.

\begin{figure}[!h]
  \centering
  \includegraphics[width=1.0
 \linewidth]{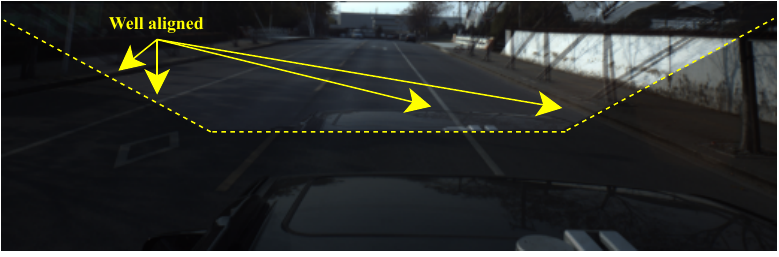}
  \caption{Correctly fused uphill frames using estimated normal.
  }
  \label{fig:fuse_uphill}
\end{figure}

Additionally, we investigate whether our method successfully estimates the road surface normal vector under uphill/downhill road scenarios. Fig.~\ref{fig:fuse_uphill} demonstrates that the normal is accurately estimated in uphill scenarios, resulting in good alignment of road marks across multiple frames.

\section{Dataset Selection Criteria}
Our method relies on borrowing information from adjacent frames, and therefore datasets that do not include continuous frames with a common field of view, such as CeyMo \cite{jayasinghe2022ceymo}, CULane \cite{shirke2019lane}, and VPGNet \cite{lee2017vpgnet}, are unsuitable for our purposes. These datasets do not provide the necessary information for our method to work effectively, as they lack the continuity required for the information to be meaningfully borrowed across frames.
In addition, although vehicle pose is free information that we can exploit, transitional lane mark detection datasets, such as , LLAMAS \cite{llamas2019}, SDLane \cite{Jin2022eigenlanes} and VIL-100 \cite{zhang2021vil}, do not provide camera intrinsic and extrinsic information, nor do they provide GPS/IMU data that could be used to calculate camera global extrinsic information. These datasets are unsuitable for our method because they lack the necessary information for our approach to work effectively. 
Furthermore, our method is a segmentation task, and as such, datasets that do not provide segmentation masks, such as OpenLane \cite{chen2022persformer}, Tusimple \cite{shirke2019lane} are also unsuitable. Without segmentation masks, it is impossible to accurately determine the boundaries of lane markings in the image, making it difficult to apply our method effectively.
While the Waymo Open Dataset \cite{sun2020scalability} primarily targets general panoptic segmentation, our lane marker segmentation task focuses on only two categories—lane markers and road markers—limiting its utility. Our comparison solely involves CFFM \cite{sun2022vss} on this dataset, highlighting our approach's enhanced efficacy in Table \ref{Tab:waymo}.

\begin{table}[h]
\caption{Comparison on Waymo Open Dataset
}
\label{Tab:waymo}
\centering
\renewcommand{\arraystretch}{1.}
\begin{tabular}{c|@{\extracolsep{\fill}}ccc}
\hline
 &  CFFM \cite{sun2022vss} & Ours \\
\hline
 3 mIoU$\uparrow$ & 63.22 & \textbf{66.45} \\
\hline
\end{tabular}
\end{table}

Given these limitations, our primary experimentation is carried out on the ApolloScape dataset \cite{huang2018apolloscape}, as it offers the necessary information for our approach. In addition, we created an artificial dataset called ApolloScape Night from the ApolloScape dataset using a cross-domain generation network \cite{arruda2019cross}. This dataset allows us to evaluate the effectiveness of our method under challenging lighting conditions.
We conducted experiments on these datasets to demonstrate the effectiveness of our method across various domains, including daytime and nighttime driving scenarios. The results show that our approach achieves SOTA performance on these datasets, confirming its effectiveness and suitability for real-world lane detection applications.

\section{Visualization on Real Night/Rain Scenes}
\begin{figure}[t]
    \centering
    \includegraphics[width=0.47\textwidth]{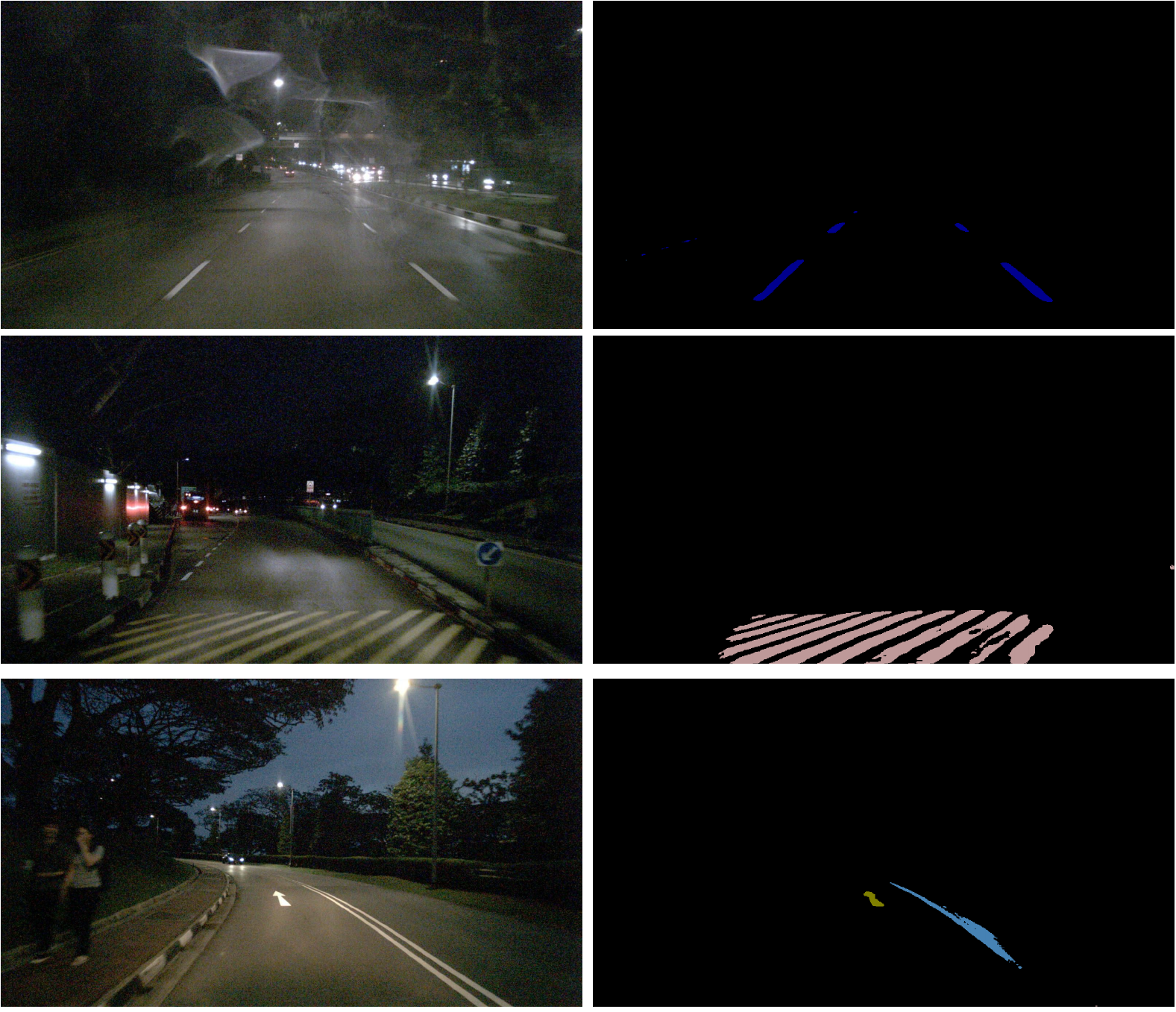}
    \caption{Segmentation results (Right) of real night frames (Left) using the model trained on the artificial ApolloScape Night dataset. Our proposed method accurately segments various road lines and markings, such as dot lines, solid lines, crosswalks, and straight arrows.}
    \label{fig:night_nu}
\end{figure}

\begin{figure}[t]
    \centering
    \includegraphics[width=0.47\textwidth]{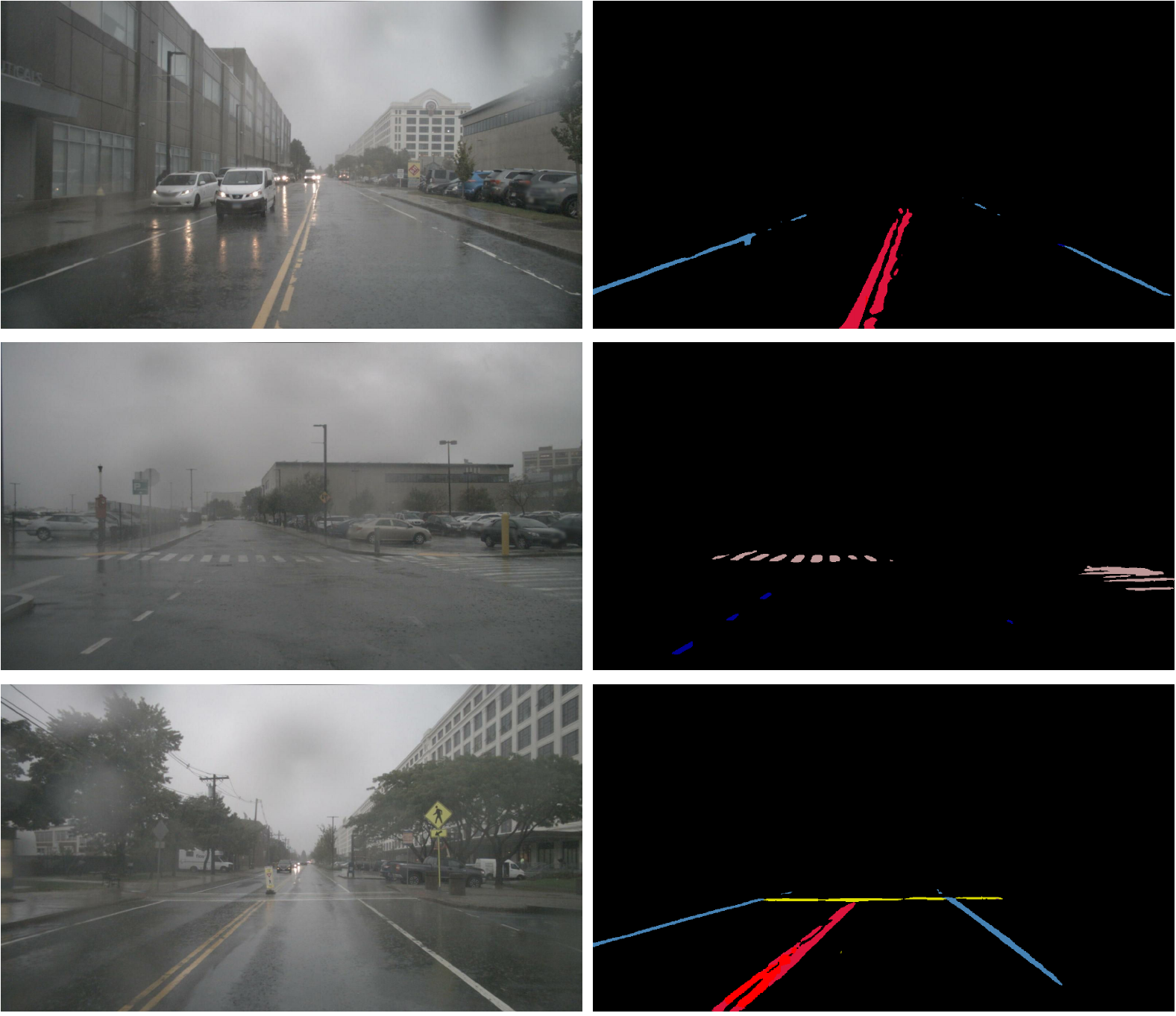}
    \caption{Segmentation results (Right) of real rainy day frames (Left) using the model trained on the artificial ApolloScape Night dataset. The proposed method accurately segments various road lines and markings, including dot lines, solid lines, double yellow lines, stop lines, and crosswalks.}
    \label{fig:rain_nu}
\end{figure}

We verify the effectiveness of our proposed method in challenging real-world   environments, \eg dark and rainy situations. Fig.~\ref{fig:night_nu} and Fig.~\ref{fig:rain_nu} present the segmentation results of our proposed method on real images taken at night and in rainy conditions, respectively. Despite being trained only on the artificial ApolloScape Night dataset, which simulates road conditions at night, our proposed method successfully segments various road lines and markings in real images. 

\begin{figure}[h]
    \centering
    \includegraphics[width=0.47\textwidth]{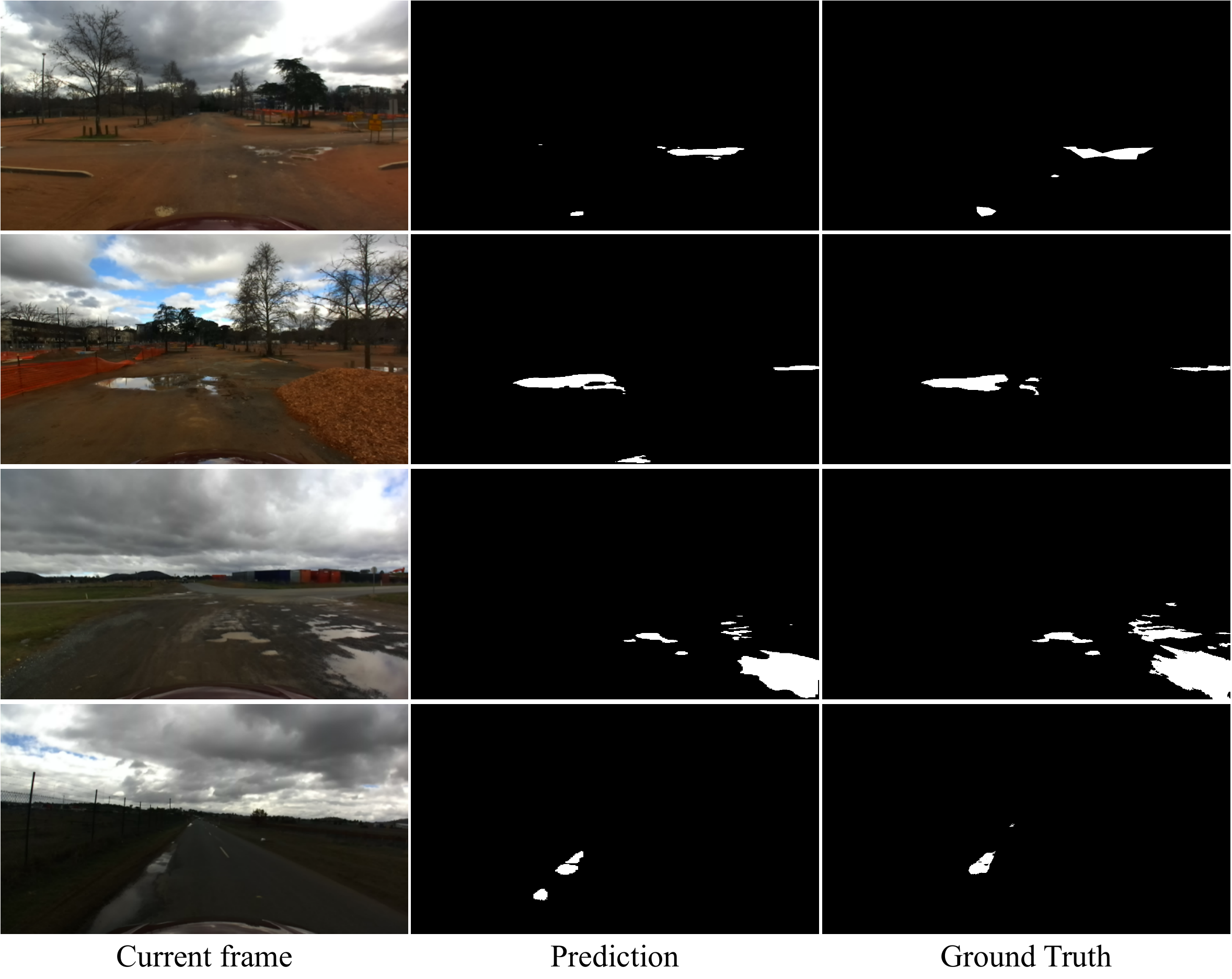}
    \caption{Current frame, prediction and ground truth masks of water puddle segmentation. The $1^{st}$ and $2^{nd}$ rows are from the Off-road dataset, and the $3^{nd}$ and $4^{th}$ rows are from the On-road dataset.}
    \label{fig:water_pred}
\end{figure}

\begin{figure}[h]
    \centering
    \includegraphics[width=0.47\textwidth]{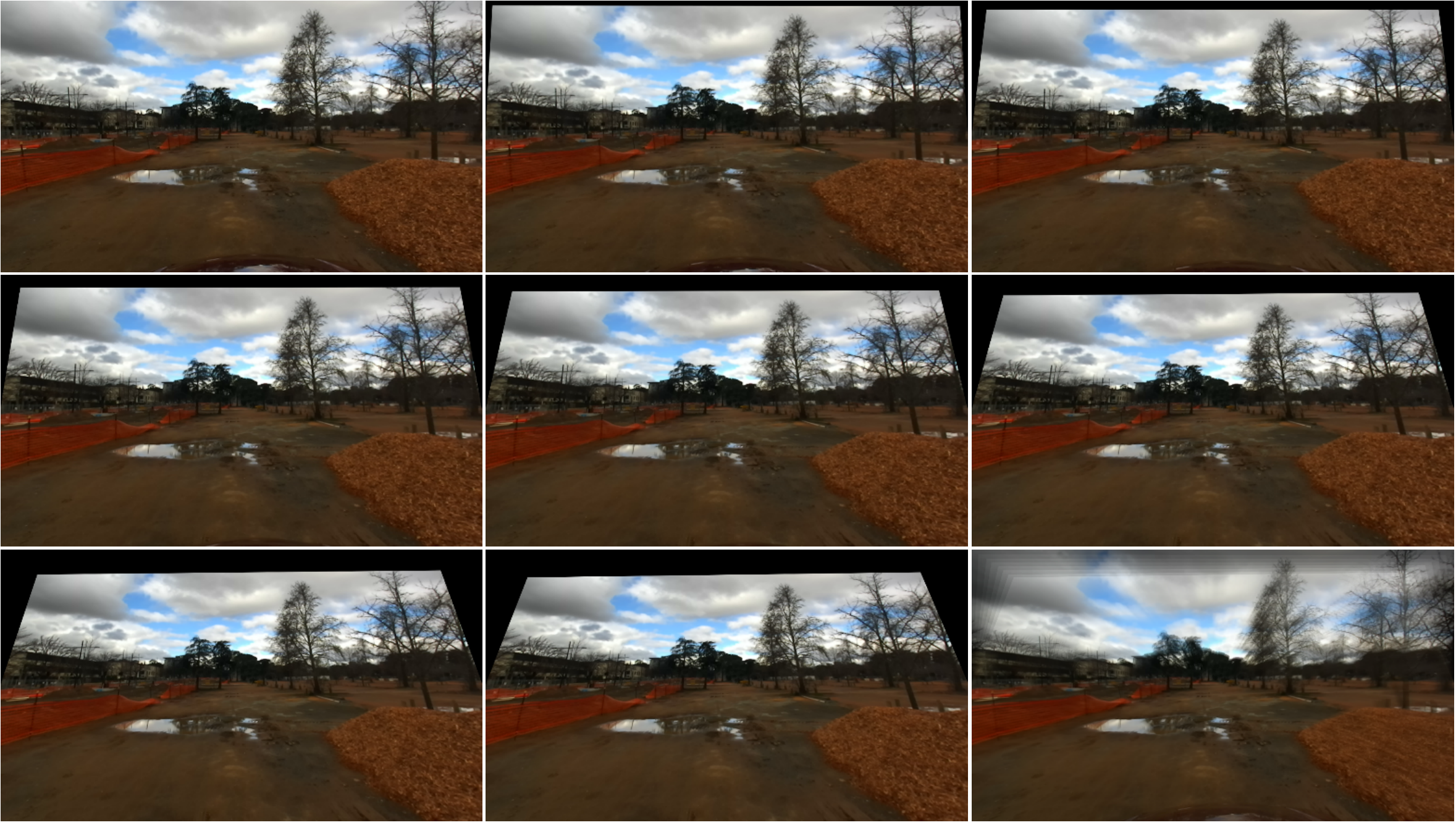}
    \caption{Input frames transformed to the current frame by homography using the estimated road surface normal vectors and camera movement. The fused image of all eight frames is shown in the bottom right corner.}
    \label{fig:water_inputs}
\end{figure}

\section{Visualization on HomoGuide}
\begin{figure*}[t]
    \centering
    \includegraphics[width=1.\textwidth]{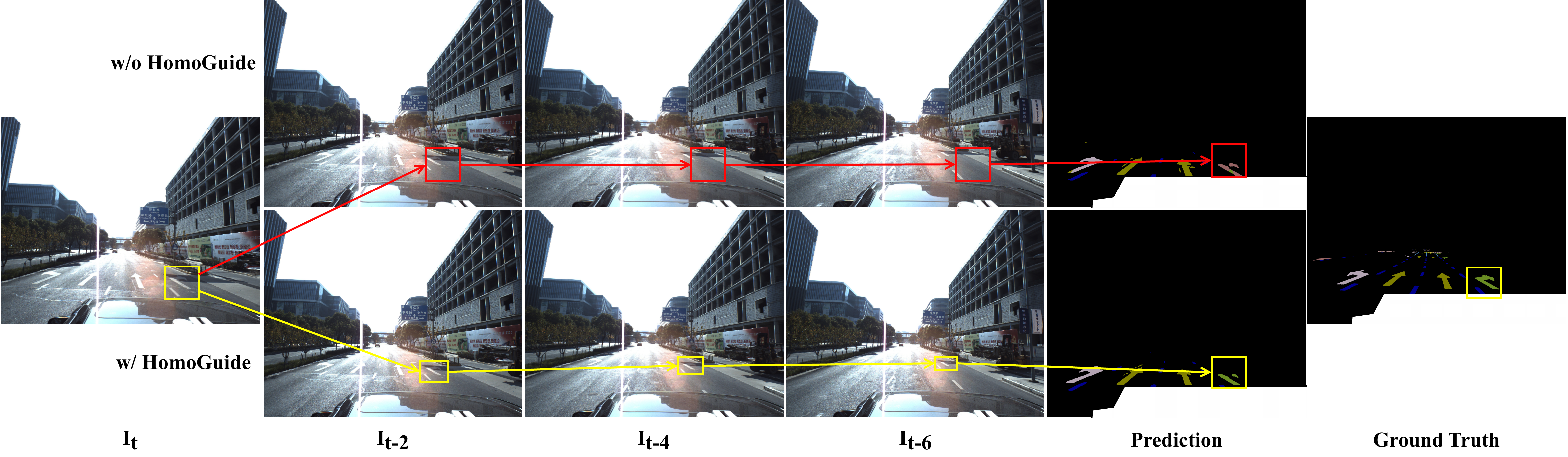}
    \caption{Impact of HomoGuide on segmentation results. Yellow boxes indicate the correspondence across frames with HomoGuide and Red boxes indicate the correspondence without HomoGuide. HomoGuide facilitates the correct classification of partially occluded road markings, as shown by the example of the partially occluded right turn arrow.
    }
    \label{fig:homog_compare}
\end{figure*}

This section complements Section.~4.6 of the Main Paper on the \enquote{Impact of HomoFusion}. 
Fig.~\ref{fig:homog_compare} compares our proposed method with and without Homo Guide. It illustrates that incorporating HomoGuide enables our method to accurately classify road lines and markings even under adverse conditions such as occlusion, road reflection, and poor light conditions.

\section{Visualization of Water Hazard Detection}
This section complements Section.~4.6 of the Main Paper on the \enquote{Application to Another Task}, we provide a visualization of the predicted segmentation and the input frames transformed by Homography in Fig.~\ref{fig:water_pred} and Fig.~\ref{fig:water_inputs}, respectively.
These figures demonstrate that our proposed HomoFusion can align ground surfaces well enough to improve the performance of detection of flat objects on the ground.

\section{Extra Qualitative Examples}
\begin{figure*}[!htb]
    \centering
    \includegraphics[width=0.99\textwidth]{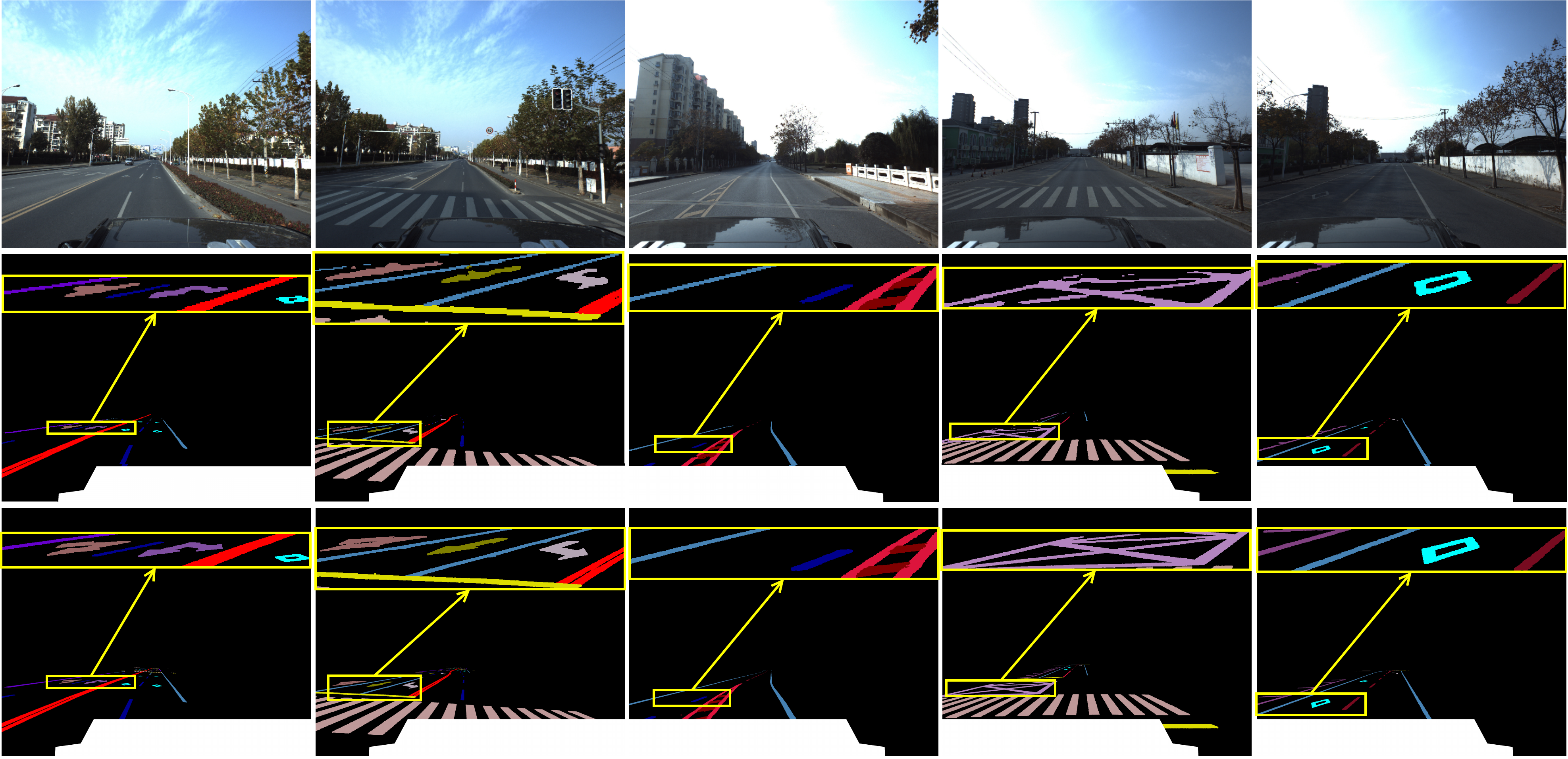}
    \caption{Examples of prediction on various road lines and markings. From top to bottom are input images, prediction of our model, and ground truth. Best viewed in color.}
    \label{fig:various_vis}
\end{figure*}

Fig.~\ref{fig:various_vis} demonstrates that our proposed method can accurately segment the road lines and markings of various categories.

\begin{figure*}[!htb]
    \centering
    \includegraphics[width=0.85\textwidth]{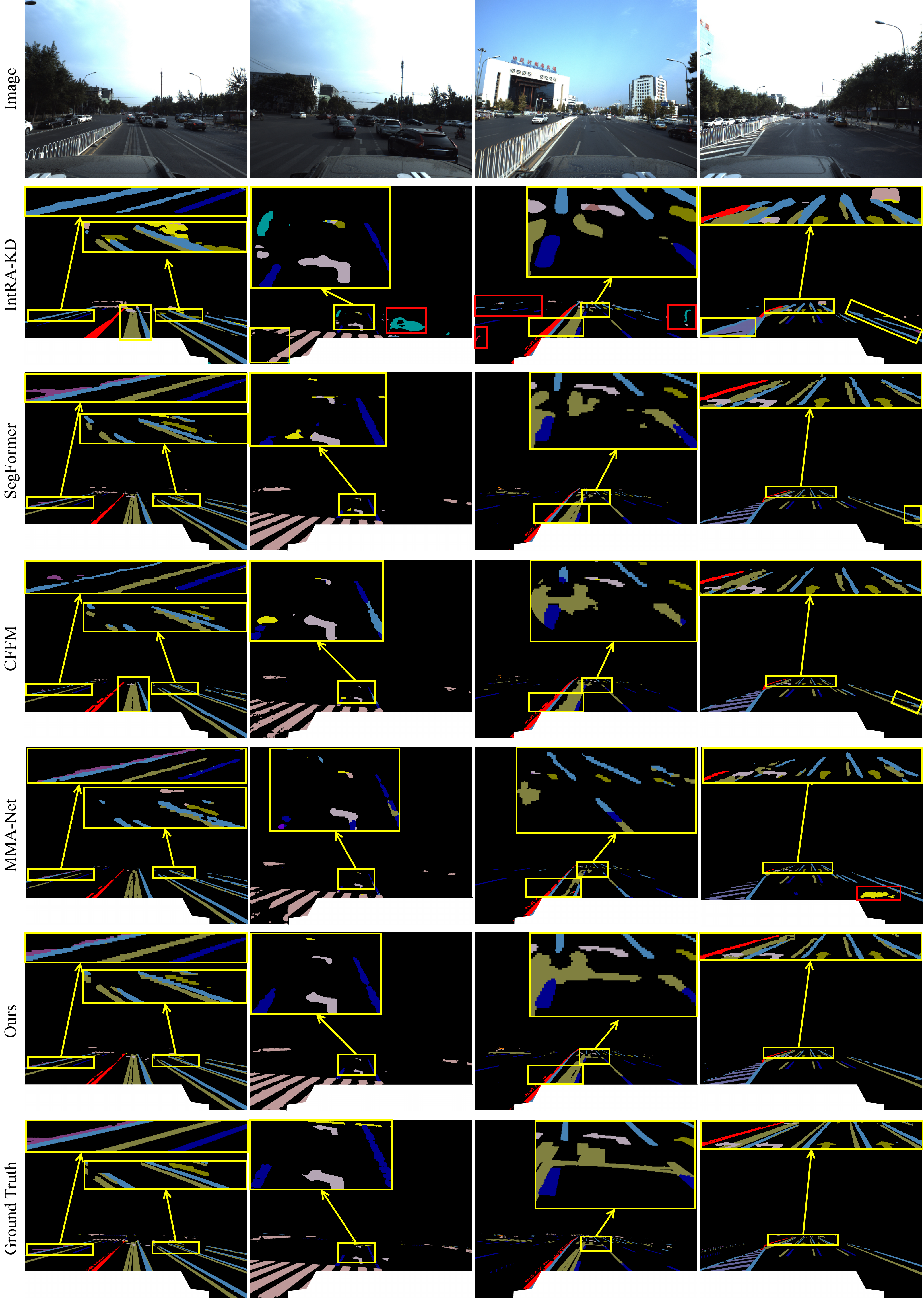}
    \caption{Extra qualitative comparison of our proposed method with SOTA methods on the ApolloScape \cite{huang2018apolloscape} dataset. Yellow boxes highlight the differences and enlarge the target area for better visibility. Red boxes indicate false-positive segmentation predictions.}
    \label{fig:extra_day}
\end{figure*}

\begin{figure*}[!htb]
    \centering
    \includegraphics[width=0.85\textwidth]{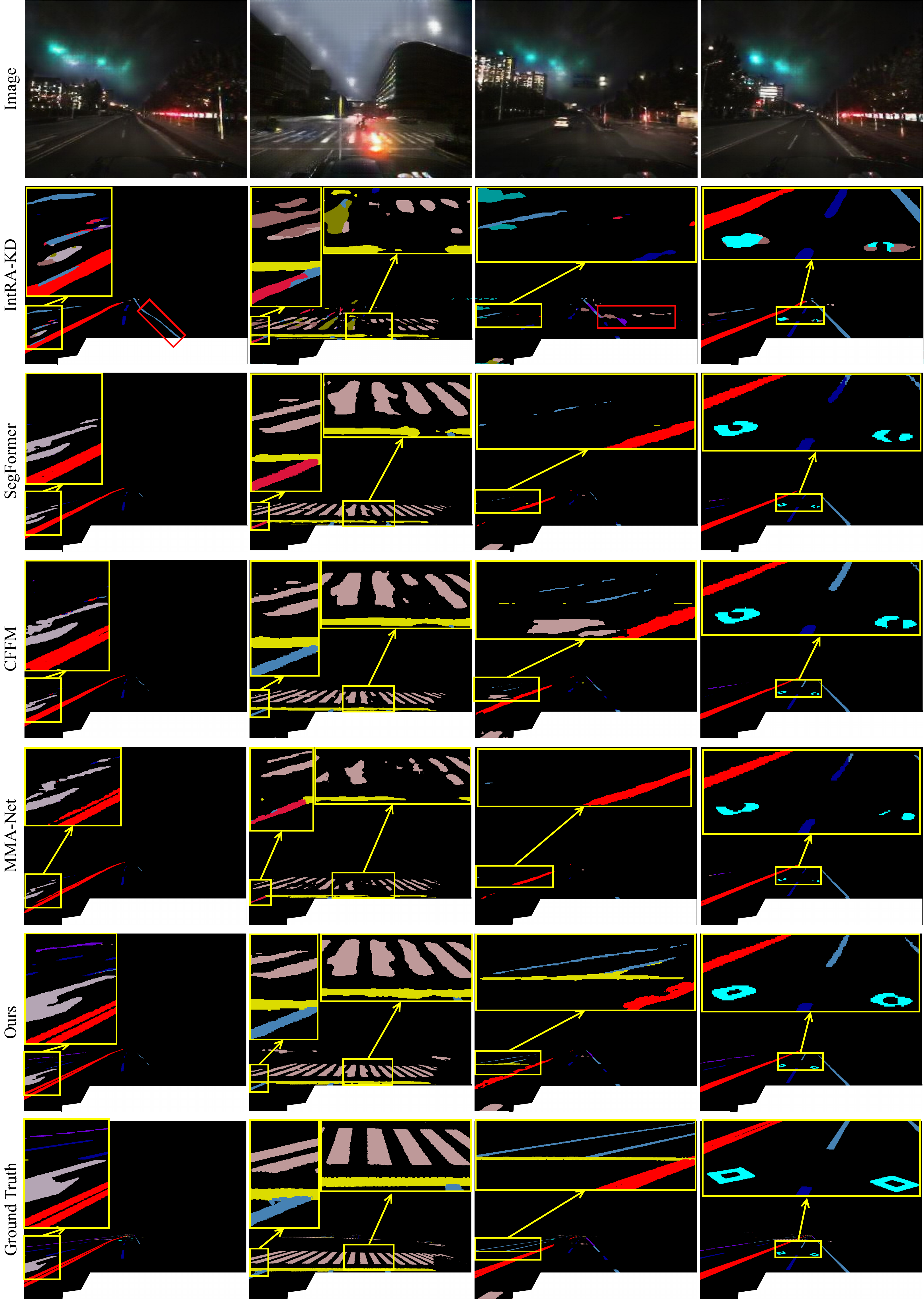}
    \caption{Extra qualitative comparison of our proposed method with SOTA methods on the ApolloScape Night dataset. Yellow boxes highlight the differences and enlarge the target area for better visibility. Red boxes indicate false-positive segmentation predictions.}
    \label{fig:extra_night}
\end{figure*}

Furthermore, we provide additional qualitative examples of our proposed method and state-of-the-art (SOTA) algorithms, including (a) IntRA-KD \cite{hou2020inter}, (b) SegFormer \cite{xie2021segformer}, and (c) CFFM \cite{sun2022vss} and (d) MMA-Net \cite{zhang2021vil}, on ApolloScape \cite{huang2018apolloscape} in Fig.\ref{fig:extra_day} and ApolloScape Night in Fig.\ref{fig:extra_night}. Our method demonstrates better true-positive predictions and fewer false-positive predictions in both scenarios. Additionally, our method accurately predicts more precise boundaries even in challenging glare and poor lighting conditions.

